\RequirePackage{fix-cm}
\documentclass[twocolumn]{article}          
\usepackage{times}
\usepackage{epsfig}
\usepackage{graphicx}
\usepackage{amsmath}
\usepackage{amssymb}
\usepackage{authblk}
\usepackage{adjustbox}

\usepackage{epstopdf}
\usepackage{subfigure}
\usepackage{caption}
\usepackage{tabularx}
\usepackage[table]{xcolor}
 \usepackage[normalem]{ulem}
\usepackage{algorithm}
\usepackage[noend]{algpseudocode}
\usepackage[section]{placeins}
\usepackage{booktabs}
\usepackage{multirow}
\usepackage{array}
\newcolumntype{P}[1]{>{\centering\arraybackslash}p{#1}}
\newcolumntype{M}[1]{>{\centering\arraybackslash}m{#1}}
\usepackage{arydshln}
\usepackage{hyperref}

\makeatletter
\def\BState{\State\hskip-\ALG@thistlm}
\makeatother
\definecolor{myGreen}{HTML}{33FF00}
\definecolor{myRed}{HTML}{FF3030}
\definecolor{myGrey}{HTML}{AA5555}
\definecolor{myWhite}{HTML}{FFFFFF}
\definecolor{maroon}{cmyk}{0,0.87,0.68,0.32}
\definecolor{petr}{HTML}{5555FF}
\definecolor{josef}{HTML}{FF3030}

\usepackage{etoolbox}

\begin{document}

\title{ReshapeGAN: Object Reshaping by Providing A Single Reference Image}



\author[1]{Ziqiang Zheng\thanks{equal contribution} }
\author[2]{Yang Wu\thanks{equal contribution}}
\author[1]{Zhibin Yu}
\author[3]{Yang Yang}
\author[1]{Haiyong Zheng}
\author[4]{Takeo Kanade}

\affil[1]{Ocean University of China.}
\affil[2]{Nara Institute of Science and Technology, Japan.}
\affil[3]{University of Electronic Science and Technology of China.}
\affil[4]{Carnegie Mellon University, United States.}

\vspace{-5mm}
\date{}
\maketitle

\begin{abstract}
The aim of this work is learning to reshape the object in an input image to an arbitrary new shape, by just simply providing a single reference image with an object instance in the desired shape. We propose a new Generative Adversarial Network (GAN) architecture for such an object reshaping problem, named ReshapeGAN. The network can be tailored for handling all kinds of problem settings, including both within-domain (or single-dataset) reshaping and cross-domain (typically across mutiple datasets) reshaping, with paired or unpaired training data. The appearance of the input object is preserved in all cases, and thus it is still identifiable after reshaping, which has never been achieved as far as we are aware. We present the tailored models of the proposed ReshapeGAN for all the problem settings, and have them tested on 8 kinds of reshaping tasks with 13 different datasets, demonstrating the ability of ReshapeGAN on generating convincing and superior results for object reshaping. To the best of our knowledge, we are the first to be able to make one GAN framework work on all such object reshaping tasks, especially the cross-domain tasks on handling multiple diverse datasets. We present here both ablation studies on our proposed ReshapeGAN models and comparisons with the state-of-the-art models when they are made comparable, using all kinds of applicable metrics that we are aware of. Code will be available at \url{https://github.com/zhengziqiang/ReshapeGAN}.



\end{abstract}


\begin{figure}[!ht]
\centering
\includegraphics[width=1.0\linewidth]{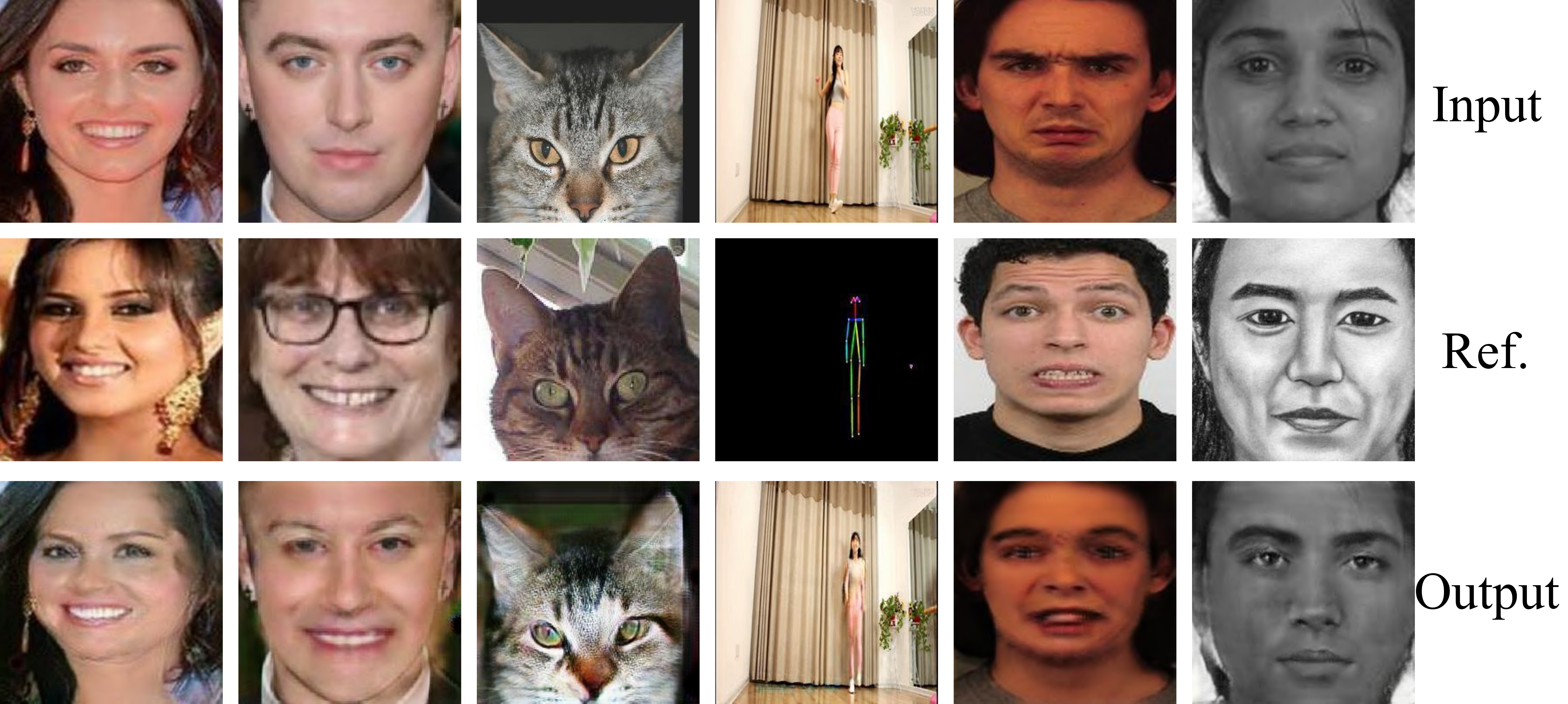}
\caption{Object reshaping with ReshapeGAN, guided by another object from a single reference image (Ref.). The appearance is preserved (thus being still identifiable) while the reshaping is controllable by just choosing a reference image with the desired shape. The reference can be from an arbitrary domain (same as or different from that of the input).}
\label{fig:problem_setting}
\end{figure}

\section{Introduction}
\label{intro}

Many of us may have admired some others' faces or bodies, though it may be hard for us to even dream about changing our own faces/bodies to those desired ones, not to say taking actions for actual reshaping. But what if the reshaping can be seen instantly, as shown in Fig.~\ref{fig:problem_setting}, by just providing a picture of ours and another one of some other person who has the desired shape? In greater details, if the virtual reshaping can preserve our identity and appearance, while at the same time transferring to the same nice expression or pose as that of the admired person, won't you want to try and check this visible dream?

\begin{figure*}[!ht]
\centering
\includegraphics[width=1.0\linewidth]{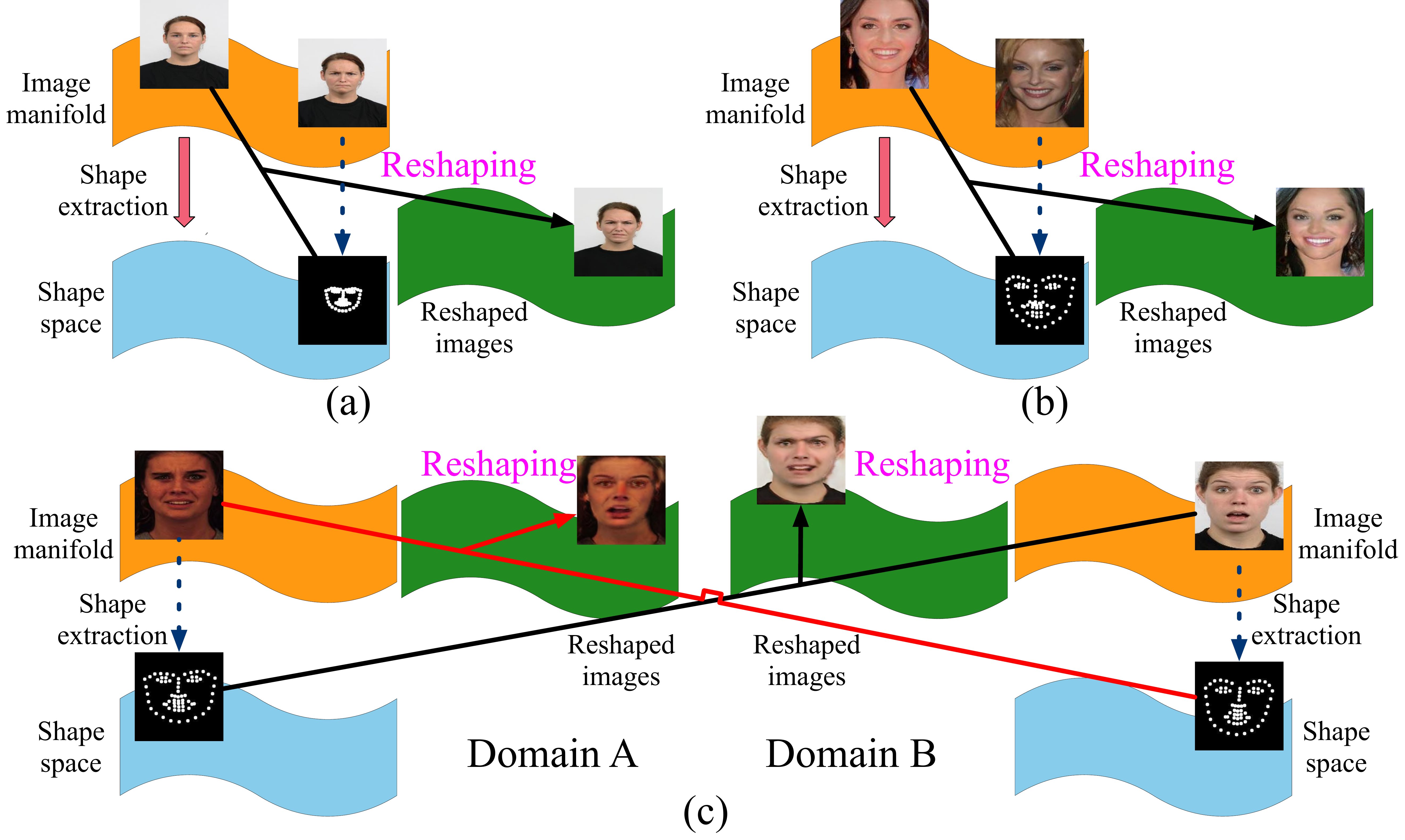}
\caption{The typical settings that ReshapeGAN handles: (a) reshaping by within-domain shape guidance with paired data, (b) reshaping by within-domain shape guidance with unpaired data, and (c) reshaping by cross-domain shape guidance.}
\label{fig:architecture}
\end{figure*}

It is actually even beyond what dreaming can do, because as humans we can hardly imagine the details of the changes caused by such a reshaping if we have never seen similar results before. A skilled artist may be able to do such translation and imagination, but it is definitely not an easy task. Nevertheless, we have the ability to feel how good such reshaping results look like, even without a ground truth. We can be somehow sensitive about how well the identity and appearance are preserved, how well the changed shape matches that of the reference image, and how realistic the generated image is. Clearly, it is fun to try such a task. 

In this paper, we propose a framework called \textbf{ReshapeGAN} that can automatically learn to do such a reshaping, of which some exemplar results are shown in Fig.~\ref{fig:problem_setting}. ReshapeGAN not only works on faces/bodies, but also applies to other objects like cats. 

ReshapeGAN is a type of Generative Adversarial Network (GAN), which was first introduced by Goodfellow et. al. in 2014~\cite{goodfellow2014generative}. GAN has been developed and proved to be very effective on image generation tasks for many applications~\cite{mirza2014conditional,zhao2016energy,shrivastava2017learning,zhang2017stackgan,zhu2017unpaired,kim2017learning,yi2017dualgan,press2017language,choi2018stargan,huang2018multimodal,lee2018diverse,lucic2018gans,kurach2018gan}. Generally, conditionally generating images could fall into two categories: \emph{supervised image generation} and \emph{unsupervised image-to-image translation}. The former can usually generate higher quality results~\cite{mirza2014conditional,yan2016attribute2image,antipov2017face}, which are more realistic~\cite{isola2017image} or with higher resolution~\cite{wang2017high}, but it has the limitation of relying on paired training data (input and ground truth pairs) which may be hard or impossible to collect in many applications. The latter doesn't require such paired training data and thus being more widely applicable. Though unsupervised image-to-image translation is more challenging, its merits have attracted researchers to make a lot of progresses including CycleGAN~\cite{zhu2017unpaired}, DualGAN~\cite{yi2017dualgan}, DiscoGAN~\cite{kim2017learning}, MUNIT~\cite{huang2018multimodal}, DRIT~\cite{lee2018diverse}, StarGAN~\cite{choi2018stargan}, etc~\cite{huang2018introduction}. Specially, the cross-domain or cross-dataset unsupervised image-to-image translation is most challenging, as each domain or dataset has its own style and attributes. Image generation have to manipulate and control such styles and attributes for a desired output~\cite{mirza2014conditional,zhao2016energy,zhu2017unpaired,kim2017learning,yi2017dualgan,press2017language,choi2018stargan,huang2018multimodal,lee2018diverse}. The proposed ReshapeGAN learns to preserve the appearance of the input object instance and get the shape information from the reference image for the reshaping tasks. Its framework can be tailored for both supervised image generation and unsupervised image-to-image translation, within a domain or across domains/datasets, as shown in~Fig~\ref{fig:architecture}. As far as we are aware, this is the first time that object reshaping is made possible for all these settings, especially for the cross-domain/cross-dataset setting. 

Usually, cross-domain image-to-image translation requires a large amount of training data from each individual for ensuring a reasonably good performance, due to the possibly large style and attribute differences between domains/datasets. Nevertheless, ReshapeGAN is made effective for working with relatively small amount of data from each individual, so that being as generally applicable as possible.

To demonstrate the effectiveness and superiority of ReshapeGAN, we conduct extensive experiments for each of the three main settings, resulting in totally 8 different tasks on 13 datasets. Since there is no unique metric that is widely recognized as the standard for generated image's quality assessment, we adopt all the applicable metrics that we are aware of and use them for performance evaluation and comparison. We compare ReshapeGAN with state-of-the-art methods on all the tasks which they can be applied to or made applicable for, and do ablation studies to verify the effectiveness of each component of ReshapeGAN.

In brief, the main contributions of this work can be summarized as follows.
\begin{itemize}
	\item It presents as far as we know the first general reference-guided object reshaping framework, which preserves the object's identity and works on most diverse input-reference pairs.  
	\item It introduces tailored models of the proposed framework, which are applicable to both supervised and unsupervised reshaping, for both within-domain and cross-domain or cross-dataset scenarios. 
	\item Extensive experiments have been done on all the settings with many tasks and datasets, to show the effectiveness of the proposal with ablation studies and its significant superiority in comparison with state-of-the-art methods when they are or are made to be comparable.
\end{itemize}

\section{Related work}

\subsection{Generative Adversarial Networks (GANs)}
Thanks to the development of GAN~\cite{goodfellow2014generative}, the recent extensions of GAN have achieved a great progress in various vision tasks such as image editing and in-painting~\cite{gatys2015neural,pathak2016context,yang2017high,bau2019gan}, super resolution~\cite{ledig2017photo,chen2017face,sonderby2017amortised}, image restoration and enhancement~\cite{luo2015removing,zhang2017image}, object detection~\cite{li2017perceptual,nogue2018object} and other applications~\cite{denton2015deep,zhao2016energy,zhang2017stackgan,mao2017least,reed2016generative,nguyen2017plug}. Regarding with adversarial training, how to avoid model collapse and reduce the training instability is an open discussion~\cite{lucic2018gans,creswell2018generative}. To achieve this goal, some studies focus on designing novel network architectures and training mechanisms~\cite{chen2016infogan,odena2017conditional} while more studies tried to solve this problem by updating adversarial loss functions such as Boundary-Seeking Generative Adversarial Networks~\cite{devon2017boundary}, Wasserstein Generative Adversarial Networks (WGAN)~\cite{arjovsky2017wasserstein}, Loss-Sensitive Generative Adversarial Networks (LS-GAN)~\cite{qi2017loss}, Least Square Generative Adversarial Networks (LSGAN)~\cite{mao2017least}, etc~\cite{kurach2018gan}. In our tasks, we also include some efficient generative adversarial losses to improve the training process and yield more plausible results. We take advantage of WGAN~\cite{arjovsky2017wasserstein} and its extension WGAN-GP~\cite{gulrajani2017improved} to make the adversarial training process more stable. Furthermore, we use Perturbed loss from DRAGAN~\cite{kodali2017convergence} to improve the synthetic quality.

\subsection{GAN based image-to-image translation}
Benefited from the success of conditional
GANs (cGANs)~\cite{mirza2014conditional}, many researches related with GANs focused on image-to-image translation tasks. According to the data type, these tasks can be roughly divided into paired~\cite{isola2017image,wang2017high,zhu2017toward} and unpaired~\cite{zhu2017unpaired,yi2017dualgan,taigman2017unsupervised,liu2017unsupervised,kim2017learning,radford2015unsupervised} image-to-image translation. Paired image-to-image translation approaches usually require the paired images for training. They always need one or more pixel-wise restriction such as L1 norm to implement supervised learning with data pairs~\cite{isola2017image}. However, paired data are expensive and sometimes impossible to be collected. To overcome this shortage, CycleGAN, DualGAN and DiscoGAN applied a cycle consistency to translate images using unpaired images~\cite{zhu2017unpaired,kim2017learning,yi2017dualgan}. Besides, Choi et al. proposed StarGAN~\cite{choi2018stargan} that used a single model and latent code to handle one-to-many image-to-image translation tasks. Similar to ACGAN~\cite{odena2017conditional}, which trained the generator with the discriminator that combines with an auxiliary classifier, StarGAN inherited the structure and the cycle consistency loss from CycleGAN and used an additional classifier to control the synthesize attribute. Besides, combining variational autoencoders (VAE)~\cite{kingma2013auto} with GANs is another important branch to translate images between different domains~\cite{larsen2016autoencoding} with unpaired data. Fader Networks~\cite{lample2017fader} used the attribute-invariant representations, encoded by the input image, and the latent code for image reconstruction. Zhu et al. proposed BicycleGAN~\cite{zhu2017toward}, which combines VAE-GAN~\cite{larsen2016autoencoding} objects and latent regressor objects~\cite{donahue2016adversarial,dumoulin2016adversarially}
for a bijective consistency to obtain more realistic and diverse samples using paired data. These studies have advanced the development of the one-to-one image-to-image translation. However, no existing work can achieve identity-preserved object reshaping for both the case of training with paired data and that with only unpaired data across domains/datasets. 


\subsection{Conditional image editing}

Conditional generative models are widely used for image synthesize under one or more given conditions. Many studies were developed based on two pioneer works: conditional variantional Autoencoder (CVAE)~\cite{yan2016attribute2image} and conditional Generative Adversarial Network (cGAN)~\cite{mirza2014conditional}. Lassner et al. proposed a conditional architecture hybridizing VAE with adversarial training to generated full-body people in clothing~\cite{lassner2017generative}. Reed et al.~\cite{reed2016generative} presented a generative model to generate bird and flower images conditioned on text descriptions by adding textual information to both generator and discriminator. They further discussed the feasibility to control the features, structure and the locations of the generated images with different conditional text-to-image models~\cite{reed2016learning,reed2016generating}. StackGAN~\cite{zhang2017stackgan} and its extension StackGAN++~\cite{zhang2018stackgan++} can also use text descriptions to generate high-quality photo-realistic images. As mentioned before, StarGAN~\cite{choi2018stargan} used one-hot encoded biases as conditions to translate images from one domain to another with unpaired data. 

Comparing with these works which use labels and texts as the conditions for image generation, using multiple images as conditions for image synthesize is more challenging. Ma et al.~\cite{ma2017pose} considered both pose images and person images as conditions to guide the network to generate a person image with a specified pose. Yang et al.~\cite{yang2018pose} further extended this idea to generate videos under the constraint of pose series. Zhao et al.~\cite{zhao2018multi} explored generating multi-view cloth images from only a single view input, while Ma et al.~\cite{ma2018exemplar} and Park et al.~\cite{park2019semantic} used an extra image as exemplar for semantic-preserved unsupervised translation, which have a similar motivation to our task. However, most of image guided image editing approaches rely on paired images to implement a pixel-wise supervised training, or limit to some low-level (color, texture, etc.) translation applications. Unlike those methods, our proposed ReshapeGAN can work on unpaired training data for high-level object reshaping tasks. We achieve this by making use of shape and appearance information in a more efficient and flexible way. Our model can generate a desired image that preserves the appearance of a specific object instance while borrowing the shape information from another image, even across domains. 


\section{Network Architecture for Object Reshaping}
\label{sec:method}


Given an input image $x \in X$ and a reference image $y \in Y$ for geometric (shape) guidance represented by $s_y$, object reshaping targets at learning a generator $G$ which can generate a new image $G(x, s_y)$ inheriting $x$'s appearance while at the same time changing to $y$'s shape $s_y$. 
Our proposal for object reshaping is called ReshapeGAN, which can be tailored for three typical settings (Fig.~\ref{fig:architecture}). We introduce each of the settings and our corresponding tailored model in the following subsections.

\subsection{Reshaping by within-domain guidance with paired data}
\label{subsec:paired}

\begin{figure}[!ht]
\centering
\includegraphics[width=1.0\linewidth]{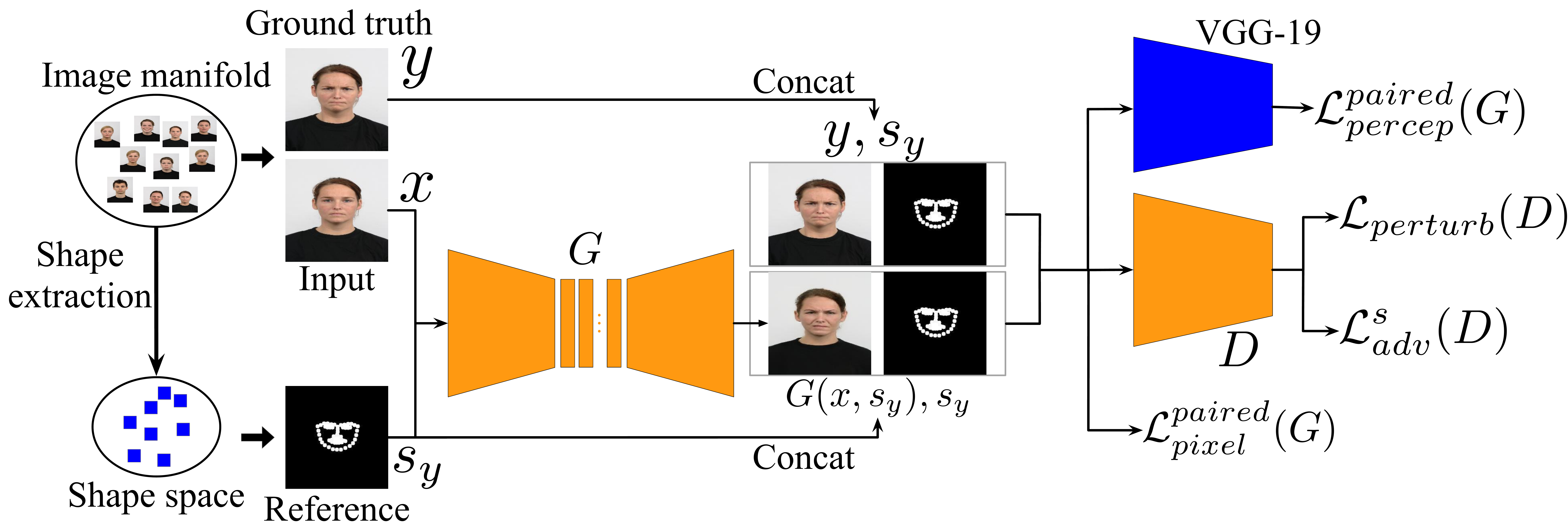}
\caption{The detailed ReshapeGAN model for reshaping by within-domain guidance with paired data.}
\label{fig:a}
\end{figure}

When the reference image is in the same domain (which usually means the same dataset or style) as the input image and the reference image is about the same object instance (i.e., having the same identity) as the one in the input image, we get the easiest setting for object reshaping, which has paired training data. In this case, ReshapeGAN can be learned by solving the following problem:
\begin{equation}
G^{*} = \arg\underset{G}{\min}\;\underset{D}{\max} \mathcal{L}^{paired}_{ReshapeGAN}(G,D),
\end{equation}
with
\begin{equation}
\begin{aligned}
\mathcal{L}^{paired}_{ReshapeGAN}(G,D)
& = \mathcal{L}^s_{adv}(G,D) + \gamma \mathcal{L}_{perturb}(D) \\
& + \delta \mathcal{L}^{paired}_{pixel}(G) + \sigma \mathcal{L}^{paired}_{percep}(G),
\end{aligned}
\end{equation}
where $D$ is its discriminator; $\mathcal{L}^s_{adv}(G,D)$ is the adversarial loss with shape guidance; $\mathcal{L}_{perturb}(D)$ is the perturbed loss for regularizing $D$; $\mathcal{L}^{paired}_{pixel}(G)$ is the paired pixel-level appearance matching loss; $\mathcal{L}^{paired}_{percep}(G)$ denotes the perceptual loss; $\gamma$, $\delta$, and $\sigma$ are super-parameters for balancing the corresponding losses. Details of the loss functions are explained as follows and the overall model is illustrated in Fig.~\ref{fig:a}.



\noindent \textbf{Shape guided adversarial loss $\mathcal{L}^s_{adv}(G,D)$.} 
Based on our shape guided generator $G$ and following the common definition of adversarial loss, we have $\mathcal{L}^s_{adv}(G,D)$ defined as 
\begin{equation}
\label{loss:adv}
\begin{aligned}
\mathcal{L}^s_{adv}(G,D) = & \mathbb{E}_{y}\left[ \log D\left(y, s_y\right)\right] \\
& + \mathbb{E}_{x,y}\left[log(1-D(G(x,s_y), s_y) \right].
\end{aligned}
\end{equation} 
The adversarial loss is a basic requirement in an adversarial network, which makes sure that the generated image $G(x,s_y)$ cannot be distinguished from the reference image $y$ by the discriminator $D$, in terms of the shape information, but not the appearance as in existing works.

\noindent \textbf{Perturbed loss $\mathcal{L}_{perturb}(D)$.}
To further improve the robustness of the discriminator, we introduce to our model the perturbed loss originated from Gulrajani et al.'s DRAGAN work~\cite{kodali2017convergence}. The perturbed loss can be expressed as
\begin{equation}
\begin{aligned}
\mathcal{L}_{perturb}(D) & = \mathbb{E}_{\hat{x},y} \left[(\left \| \bigtriangledown_{\hat{x}}D(\hat{x}, s_y)\right \| _{2}- 1)^{2}\right], \\
\text{with} \ \ \hat{x} & =(1-\alpha)x+\alpha z, \text{where}\ \ \alpha\sim U[0, 1].
\end{aligned}
\end{equation}
Here $x$ represents a real image sample and $z$ means the random noise, which follows the Gaussian distribution; $\alpha$ is the random hyper-parameter that controls the balance between real and noise and follows the continuous uniform distribution between 0 and 1; $\bigtriangledown_{\hat{x}}$ is the gradient of $D(\hat{x},s_y)$. Please note that $\hat{x}$ is only used for calculating $\mathcal{L}_{perturb}(D)$. We believe that such a perturbed loss makes the convergence more stable and generate images with higher quality. The experiment results in Section~\ref{sec:exp} also support this point.

\noindent \textbf{Paired pixel-level appearance matching loss $\mathcal{L}_{pixel}(G)$.}
It is a widely used loss for ensuring that the generated image matches the ground truth image or has a desire appearance at the pixel-level, and usually the L1 norm is used. When paired training data is available, the loss is simple as:
\begin{equation}
\label{eq:pixel-loss_paired}
\mathcal{L}^{paired}_{pixel}(G) = \mathbb{E}_{x,y}\left[ \left\| G(x,s_y)-y\right\|_{1}\right],
\end{equation}
which is about the generation loss w.r.t. the provided ground truth data $y$.

\noindent \textbf{Paired perceptual loss $\mathcal{L}^{paired}_{percep}(G)$.}
Unlike the pixel-level loss which matches two images pixel by pixel, the perceptual loss measures the similarly at the feature level. In greater details, we follow Chen et al.'s work~\cite{chen2017photographic} and include an extra pre-trained VGG-19 network on ImageNet to compute the perceptual loss~\cite{ledig2017photo,dosovitskiy2016generating,chen2017photographic}. Compared to pixel-level loss, this loss combines the distance at multiple scales of feature representation, which could carry low-level and high-level information of images. The perceptual loss is described as:
\begin{equation}
\label{eq:paired_perceptual_loss}
\mathcal{L}^{paired}_{percep}(G) =\sum^{N}_{n}{\lambda}_{n}\mathbb{E}_{x,y}\left[||{\Phi}_{n}(y)-{\Phi}_{n}(G(x,s_y))||_{1}\right],
\end{equation}
where ${\Phi}_{n}$ is the feature extractor at the $n_{th}$ level of the pre-trained VGG-19 network. Following~\cite{chen2017photographic}, we compute the perceptual loss between outputs at defined $N=5$ selective layers. The weights of pre-trained model will not be optimized during the learning process. The hyper-parameter ${\lambda}_{n}$ controls the influence of perceptual loss at different scales. The perceptual loss from the higher layer controls the global structure, and the loss from the low layer controls the local details during generation. Thus, the generator should provide better synthesized images to cheat this hybrid discriminator and finally improve the synthesized image quality. Please note, only for the paired training, we compute the perceptual loss between the target images and generated images. The detail information of the proposed method for reshaping by within-domain guidance with paired data could be found in Fig.~\ref{fig:a}.


\subsection{Reshaping by within-domain guidance with unpaired data}
\label{subsec:unpaired}

\begin{figure}[!ht]
\centering
\includegraphics[width=1.0\linewidth]{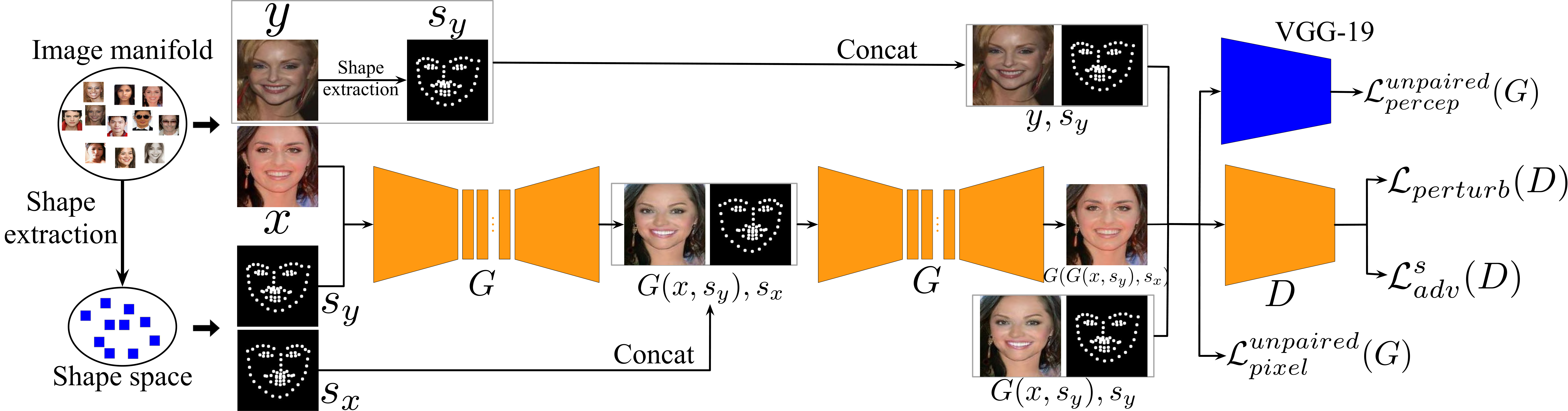}
\caption{The detailed ReshapeGAN model for reshaping by within-domain guidance with unpaired data.}
\label{fig:b}
\end{figure}

For the case that the reference image is in the same domain as the input image but they are not about the same instance/identity, the model has to be learned with unpaired training data. In this case, we use a model similar to the one introduced in Section~\ref{subsec:paired}, but having the pixel-level appearance matching loss and perceptual loss in their unpaired version. In great detail, the overall loss function becomes
\begin{equation}
\label{eq:overall_loss_unpaired}
\begin{aligned}
\mathcal{L}^{unpaired}_{ReshapeGAN}(G,D)
& = \mathcal{L}^s_{adv}(G,D) + \gamma \mathcal{L}_{perturb}(D) \\
& + \delta \mathcal{L}^{unpaired}_{pixel}(G) + \sigma \mathcal{L}^{unpaired}_{percep}(G),
\end{aligned}
\end{equation}
where the unpaired version of pixel-level appearance matching loss $\mathcal{L}^{unpaired}_{pixel}(G)$ and perceptual loss $\mathcal{L}^{unpaired}_{percep}(G)$ are defined below, and this ReshapeGAN model is illustrated in Fig.~\ref{fig:b}.

\noindent \textbf{Unpaired pixel-level appearance matching loss $\mathcal{L}^{unpaired}_{pixel}(G)$.}
When there is no paired data for training, inspired by CycleGAN~\cite{zhu2017unpaired}, we use the cycle consistency reconstruction loss with two rounds of reshaping guided by $s_x$ and $s_y$ (shapes of $x$ and $y$, respectively) as the unpaired pixel-level appearance matching loss:
\begin{equation}
\label{eq:pixel-loss_unpaired}
\mathcal{L}^{unpaired}_{pixel}(G) = \mathbb{E}_{x,y}\left[ \left\| G(G(x,s_y),s_x)-x\right\|_{1}\right].
\end{equation}

\noindent \textbf{Unpaired perceptual loss $\mathcal{L}^{unpaired}_{percep}(G)$.}
In the case that there is no paired sample with the given input $x$, it is reasonable to assume that the generated image $G(x,s_y)$ has the same perceptual response as that of $x$ due to the desired appearance and identity preserving property, those they shall have different shapes ($s_y$ vs. $s_x$). Unlike the pixel-wise matching loss which requires good alignment and thus minimum shape difference, perceptual loss is about feature-level perception results, which can have some robustness to shape changes. Therefore, we define the unpaired perceptual loss to be the that between the generated image $G(x,s_y)$ and the input image $x$:
\begin{equation}
\mathcal{L}^{unpaired}_{percep}(G) =\sum^{N}_{n}{\lambda}_{n}\mathbb{E}_{x,y}\left[||{\Phi}_{n}(x)-{\Phi}_{n}(G(x,s_y))||_{1}\right],
\end{equation}
where ${\Phi}_{n}$ is the feature extractor at the $n_{th}$ level of the pre-trained VGG-19 network, the same as the one in Equation~\ref{eq:paired_perceptual_loss}.

\subsection{Reshaping by cross-domain guidance}

\begin{figure*}[!ht]
\centering
\includegraphics[width=1.0\linewidth]{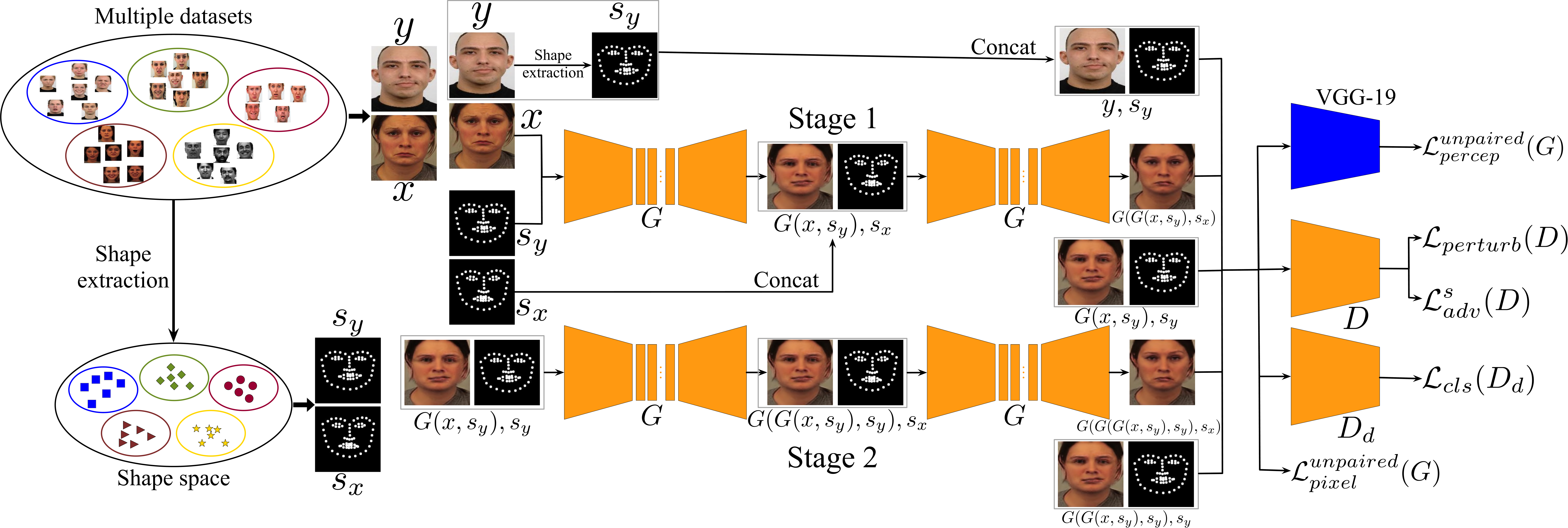}
\caption{The detailed ReshapeGAN model for reshaping by cross-domain guidance with unpaired data.}
\label{fig:c}
\end{figure*}

The hardest setting for object reshaping is about cross-domain guidance, i.e., the case when the reference image and input images are from two different domains (e.g. two different styles or datasets). For this setting, we found that it is hard for the unpaired ReshapeGAN model introduced in Section~\ref{subsec:unpaired} to learn stable appearance preserved object reshaping due to the possibly large domain differences. Therefore, we propose a two-stage strategy for dividing the whole task into two sub-tasks: domain-preserved reshaping and refining. The detailed framework is shown in Fig.~\ref{fig:c}. 

For the first stage, we add a \emph{domain classification loss} to the overall loss function of $\mathcal{L}^{unpaired}_{ReshapeGAN}(G,D)$ to ensure proper domain preservation during the reshaping. In greater details, the overall loss becomes
\begin{equation}
\label{eq:overall_loss_cross_domain_1}
\begin{aligned}
&\mathcal{L}^{cross-domain\_1}_{ReshapeGAN}(G,D,D_d) \\
&= \mathcal{L}^{unpaired}_{ReshapeGAN}(G,D) + \lambda \mathcal{L}_{cls}(D_d) \\
&= \mathcal{L}^s_{adv}(G,D) + \gamma \mathcal{L}_{perturb}(D) \\
& \quad + \delta \mathcal{L}^{unpaired}_{pixel}(G) +  \sigma \mathcal{L}^{unpaired}_{percep}(G) + \lambda \mathcal{L}_{cls}(D_d),
\end{aligned}
\end{equation}
where $D_d$ is a domain classifier and $\mathcal{L}_{cls}(D_d)$ is the domain classification loss, with $\lambda$ as its weight.

\noindent \textbf{Domain classification loss $\mathcal{L}_{cls}(D_d)$.}
The domain classifier $D_d$ is expected to be able to help ensuring the domain preserving, leading to the effect of appearance preserving, i.e., maintaining the appearance and identity of the object during the reshaping.
More specifically, $\mathcal{L}_{cls}(D_d)$ is defined as
\begin{equation}
\label{eq:loss_domain_classification}
\begin{aligned}
&\mathcal{L}_{cls}(D_d) \\
&=\mathbb{E}_{x}\left[-\log D_d(x,d_x)\right] + \mathbb{E}_{y}\left[-\log D_d(y,d_y)\right] \\
& \quad + \mathbb{E}_{x,y}\left[-\log D_d(G(x,s_y),d_x)\right],
\end{aligned}
\end{equation}
where $d_x$ and $d_y$ denote the domain labels of $x \in X$ and $y \in Y$, respectively. Here we derive this loss from StarGAN~\cite{choi2018stargan}, and we use the one-hot encoding to capture the domain distribution. Our strategy is similar to that in the work of MUNIT~\cite{huang2018multimodal}, which tries to learn disentangled representation to get more expressive and representative style information. Following the CRN~\cite{chen2017photographic}, we use the channel-wise concatenation to integrate the image and domain label (e.g. $x$ and $d_x$) for the classifier $D_d$.

In the second stage, we apply the object reshaping again by making use of the $\mathcal{L}^{unpaired}_{ReshapeGAN}(G,D)$, so that a second round of reshaping (should be significantly minor than that in the first stage) can be performed. We take the output of the first stage as the input of this stage. The overall loss for this stage is just
\begin{equation}
\label{eq:overall_loss_cross_domain_2}
\begin{aligned}
&\mathcal{L}^{cross-domain\_2}_{ReshapeGAN}(G,D,D_d) = \mathcal{L}^{unpaired}_{ReshapeGAN}(G,D)\\
&= \mathcal{L}^s_{adv}(G,D) + \gamma \mathcal{L}_{perturb}(D) \\
& \quad + \delta \mathcal{L}^{unpaired}_{pixel}(G) +  \sigma \mathcal{L}^{unpaired}_{percep}(G).
\end{aligned}
\end{equation}

The reason for splitting a difficult cross-domain object reshaping task to two easier sub-tasks is that we can get better performance than that of a one-stage generation (using only the first stage). In the one-stage generation (the first stage of our proposal), two types of information (domain/style information $d$ and shape/geometric information $s$) are used as constrains, making it hard to ensure quality reshaping. Ablation study will be given on verifying this in the following experiment section.


\subsection{Implementation details}
\label{subsec:implementation_details}

Our backbone network derives from StarGAN~\cite{choi2018stargan}, but different from~\cite{choi2018stargan}, we concatenate geometric guidance and raw input to get a new input for both generator and discriminator. Besides, in order to capture domain label effectively, we devise a new architecture for generator. In our experiments, we find that the model performs well when we choose $\gamma=\eta=1$ and $\delta=10$. Our final loss is the sum of these losses as described in Eq. \ref{eq:overall_loss_unpaired}, \ref{eq:overall_loss_cross_domain_1} and \ref{eq:overall_loss_cross_domain_2}. We apply this final loss to the generator with Adam optimizer of learning rate $0.0002$. Our code will be made available at \url{https://github.com/zhengziqiang/ReshapeGAN}. All detailed implementation could be found in our code.

\section{Experiments and results}
\label{sec:exp}

\subsection{Evaluation metrics}\label{sec:evaluation}

\noindent\textbf{Learned Perceptual Image Patch Similarity} (LPIPS) is first proposed by~\cite{zhang2018unreasonable}, which computes the perceptual similarity (actually in terms of distance) between two image patches. Lower LPIPS means the two image patches have higher perceptual similarity. Considering two image domains, we can compute the LPIPS metric (averaged over sampled patch pairs) to evaluate the perceptual similarity between them. 

\noindent\textbf{Fr\'echet Inception Distance} (FID) computes the similarity between the generated sample distribution and real data distribution. This metric is consistent and robust for evaluating the quality of generated images~\cite{lucic2018gans,borji2019pros}, and it can be calculated by:
\begin{equation}
\text{FID}=||\mu_x-\mu_g||_2^2+Tr\left(\textstyle\sum_{x}+\sum_{g}-2(\sum_{x}\sum_{g})^{\frac{1}{2}}\right),
\end{equation}
where $(\mu_{x},\sum_{x})$ and $(\mu_{g},\sum_{g})$ are pairs of the mean and covariance of the sample embeddings from the real data distribution and generated data distribution, respectively. Lower FID means smaller distribution difference between the generated and the target images and therefore higher quality of generated images.

\noindent\textbf{Geometrical Consistency} is required for our object reshaping tasks to evaluate the matching level geometrically. In order to know whether the model can synthesize images with desired geometric information, we use available state-of-the-art geometry estimation model to extract the geometric information (such as landmarks and poses) from both the synthesized images and the target images, and compare the results from sample pairs in the shape space. For facial expression generation, we use dlib~\cite{king2009dlib} to extract landmarks and measure the similarity between two images by computing SSIM (see Traditional Evaluation Metrics below for details) and LPIPS scores using the landmark information, marked as \textbf{SSIM (Landmark)} and \textbf{LPIPS (Landmark)} respectively. 

\noindent\textbf{Identification Distance} is important for our object reshaping tasks to evaluate whether the appearance and identity information are preserved while the objects are reshaped. We adopt object instance recognition (i.e., identification) algorithms to evaluate the similarity or distance between generated images and input images.
In the case of faces, we use an effective open-source face recognition algorithm\footnote{\url{https://github.com/ageitgey/face_recognition}} to do that. A significantly large distance according to the classifier, i.e., more than the normal cutoff 0.6, indicates that the two faces are from two different identities. That is, if the distance between two faces is lower than 0.6, we can consider that the two faces have the same identity. Moreover, theoretically, for the distance below 0.6, larger distance might indicate better reshaping with preserved identity, while too small distance may mean that the model fails to do reshaping so that the generated image is almost identical to the input image. This can also be proved by computing the identification distances on RaFD~\cite{rafdcite} dataset, where each identity has 8 different emotional expressions captured from 5 different angles/viewpoints ($0^\circ$, $\pm 45^\circ$ and $\pm 90^\circ$), and the average distance between the neutral faces and the other emotional expressions from $0^\circ$ viewpoint of all identities is about $0.3$, while, the average distance between the neutral faces from $0^\circ$ viewpoint and the other emotional expressions from $\pm 45^\circ$ viewpoint is about $0.5$, and the average distance between the neutral faces from $0^\circ$ viewpoint and the other emotional expressions from $\pm 90^\circ$ viewpoint is about $0.6$.

\noindent\textbf{User Study (Identity / Shape)} is the human evaluation applicable for all the generated images. We also use this golden standard for our object reshaping tasks. Here we ask 20 volunteers (users) to give a True/False judgement on the output images. Specifically, we give the user an input image, a reference image and a synthesized image, and ask them to judge true or false whether the synthesized image keep the identity information (Identity) and whether the synthesized image has the same shape expression with the reference image (Shape). The users are given unlimited time to make the decision. For each comparison, we randomly generate 100 images and each image is judged by at least 2 different users. Only higher identity metric or only higher shape metric can not provide confident performance for our object reshaping tasks, since they always require both identity preservation and reshaping ability. So higher votes to both same identity and consistent shape could indicate better reshaping performance.

\noindent\textbf{Traditional Evaluation Metrics}. For the cases where paired data exists, we can use some traditional image quality assessment metrics for performance evaluation, including \textbf{MSE}, \textbf{RMSE}, \textbf{PSNR} and \textbf{SSIM}. MSE (Mean Square Error) and RMSE (Root Mean Square Error)~\cite{willmott2005advantages} compute pixel-wise errors between synthesized images and real images, lower MSE and RMSE represent higher image generation quality.
PSNR (Peak Signal-to-Noise Ratio)~\cite{hore2010image} can roughly evaluate the image quality independently, and usually the higher PSNR the better.
SSIM (Structural Similarity Index Measure)~\cite{hore2010image} measures the similarity between two images, and higher SSIM denotes higher structural similarity between generated images and real images.

\subsection{Reshaping by within-domain guidance with paired data}

\subsubsection{Facial expression generation}\label{sec:facialexpression}

\begin{table}[!ht]
\centering
\caption{Quantitative comparison of facial expression generation on KDEF dataset. The symbol $\uparrow$ ($\downarrow$) indicates that the larger (smaller) the value, the better the performance.}
\label{table:kdef}
\begin{tabular}{@{}P{1.35cm}M{.7cm}M{.7cm}M{.7cm}M{.7cm}M{.7cm}M{1.1cm}@{}}
\toprule
Method & SSIM$\uparrow$ & PSNR$\uparrow$ & MSE$\downarrow$ & RMSE$\downarrow$ & LPIPS$\downarrow$ & FID$\downarrow$\\
\midrule
Pix2pix~\cite{isola2017image} & 0.8673 & 18.0138 & 0.0160 & 0.1133 & 0.2603 & 140.7901\\
$\text{PG}^2$~\cite{ma2017pose} & 0.9118 & 19.2469 & \textbf{0.0122} & \textbf{0.0946} & 0.2032 & 97.2435\\
ReshapeGAN & \textbf{0.9227}& \textbf{19.7374} & 0.0123 &0.0949 & \textbf{0.1808} & \textbf{84.1437}\\
\bottomrule
\end{tabular}
\end{table}

\begin{table}[!ht]
\centering
\caption{Quantitative comparison of facial expression generation on RaFD dataset. The symbol $\uparrow$ ($\downarrow$) indicates that the larger (smaller) the value, the better the performance.}
\label{table:rafd}
\begin{tabular}{@{}P{1.35cm}M{.7cm}M{.7cm}M{.7cm}M{.7cm}M{.7cm}M{1.1cm}@{}}
\toprule
Method & SSIM$\uparrow$ & PSNR$\uparrow$ & MSE$\downarrow$ & RMSE$\downarrow$ & LPIPS$\downarrow$ & FID$\downarrow$\\
\midrule
Pix2pix~\cite{isola2017image} & 0.9565 & 19.0794 & 0.0130 & 0.1121 & 0.1379 & 105.0758\\
$\text{PG}^2$~\cite{ma2017pose} & 0.9631 & 19.6109 & 0.0115 & 0.1055 & 0.1297 & 102.6674 \\
ReshapeGAN & \textbf{0.9641} & \textbf{20.1008} & \textbf{0.0103} & \textbf{0.0998} & \textbf{0.1030} & \textbf{49.2096}\\
\bottomrule
\end{tabular}
\end{table}

\begin{figure}[!ht]
\centering
\includegraphics[width=1\linewidth]{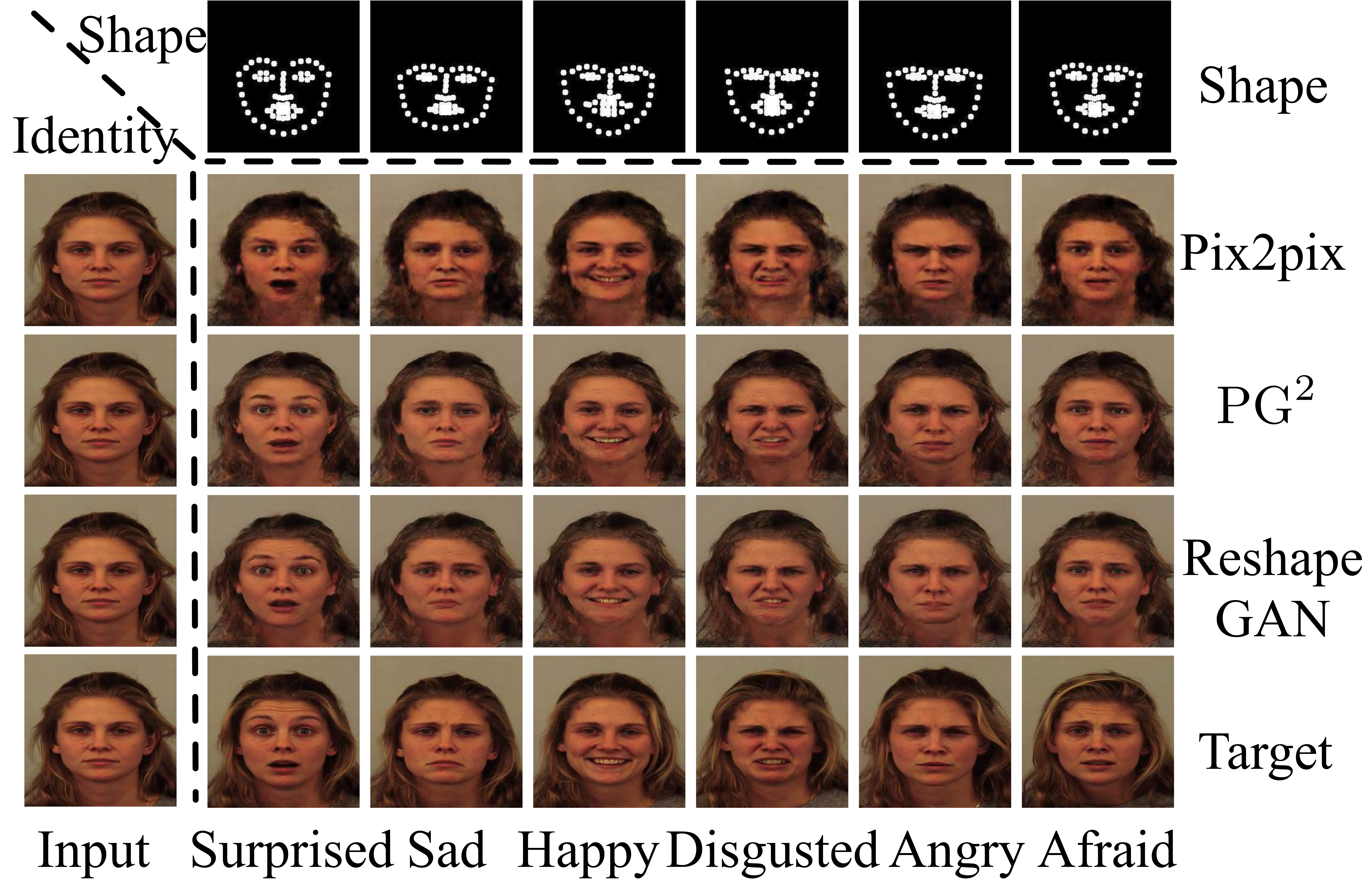}
\caption{Visual comparison of facial expression generation on KDEF dataset.}
\label{fig:kdef}
\end{figure}

\begin{figure}[!ht]
\centering
\includegraphics[width=1\linewidth]{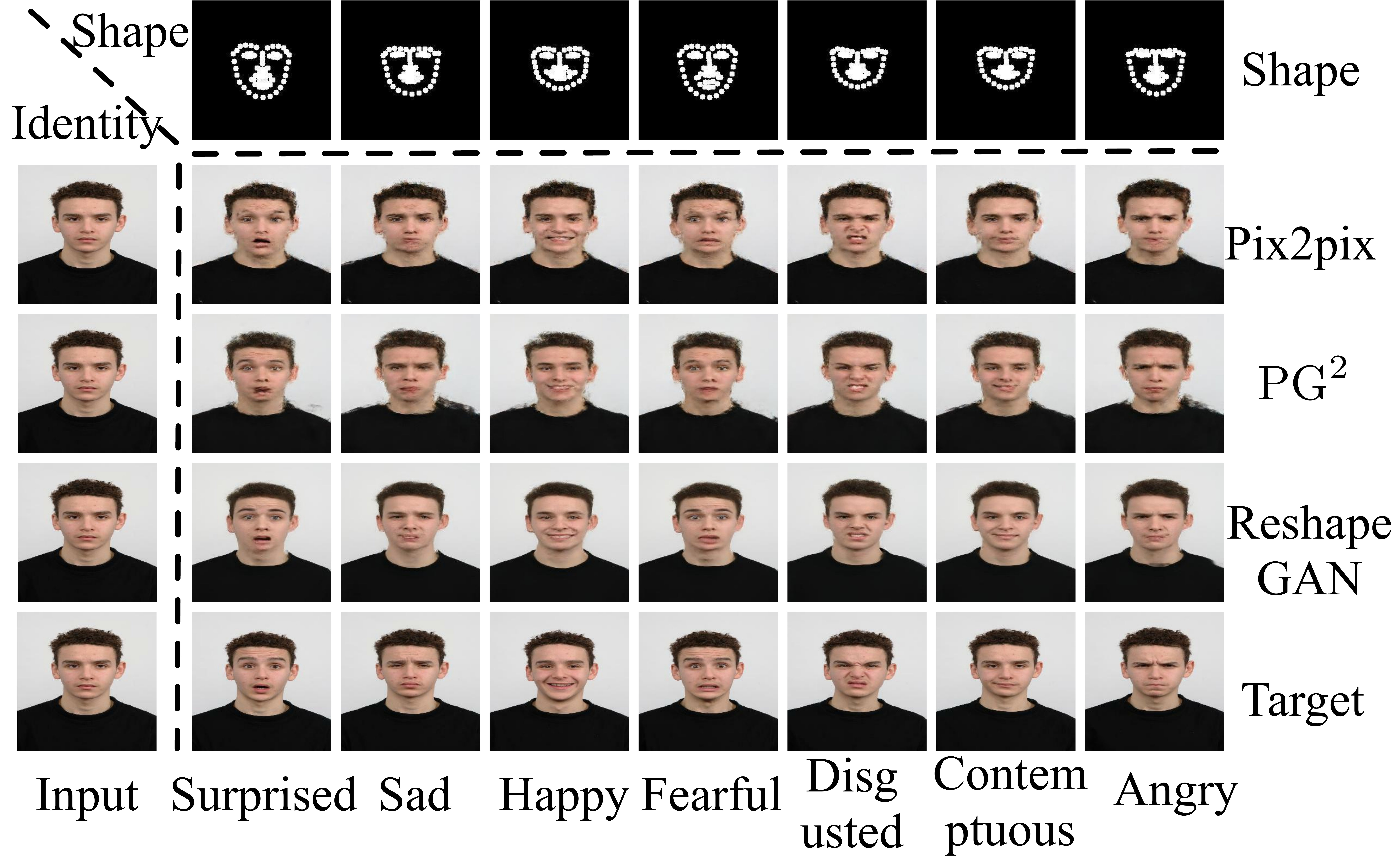}
\caption{Visual comparison of facial expression generation on RaFD dataset.}
\label{fig:rafd}
\end{figure}

First, to evaluate the image generation performance for reshaping, we conduct experiments on facial expression datasets using KDEF~\cite{calvo2008facial} and RaFD~\cite{rafdcite} datasets. For these two facial datasets, we use paired data to train all the models. KDEF contains 70 different identities with 7 different emotional representations and 5 different poses, while RaFD contains 67 identities with 8 different emotional expressions, 5 different poses and 3 eye gaze directions. For both of the two emotional datasets, we only use the frontal images with neutral faces as input images and generate images with other facial expressions. So we reorganize the two datasets, thus make KDEF (420 totally, 336 for training and 84 for evaluating) and RaFD (469 totally, 392 for training and 77 for evaluating) for facial expression reshaping. 

Here we compare our ReshapeGAN with two state-of-the-art supervised methods, i.e., Pix2pix~\cite{isola2017image} and $\text{PG}^2$~\cite{ma2017pose} for evaluation, and the quantitative comparison results can be found in Table~\ref{table:kdef} and Table~\ref{table:rafd}, in terms of SSIM, PSNR, MSE, RMSE, LPIPS and FID, where our ReshapeGAN gets the best performance among the three methods. For ReshapeGAN and $\text{PG}^2$, we use dlib~\cite{king2009dlib} to obtain the geometric information as guidance. While for Pix2pix, to compare fairly, we encode the emotional or viewpoint expression as one-hot code and inject it into the bottleneck of generator as guidance. More visual results can be found in Fig.~\ref{fig:kdef} and Fig.~\ref{fig:rafd}. As it can be seen, Pix2pix can not generate acceptable results, while the outputs of $\text{PG}^2$ have blur boundary and some dirty color blocks. Our ReshapeGAN generates reasonably clear outputs which preserve the identity information of inputs.

To investigate the efficiency of different components in our approach, we design additional experiments on RaFD for ablation study, and the quantitative results are listed in Table~\ref{table:rafd_ablation}. It can be seen that the perturbed loss $\mathcal{L}_{perturb}$ improves the PSNR score dramatically, and the geometric information helps to reduce the LPIPS distance, while the perceptual loss $\mathcal{L}_{percep}$ reduces the LPIPS and FID, since the perceptual loss actually provides multi-scale constraints to the generator by using a cascade architecture.

\begin{table}[!ht]
\centering
\caption{Quantitative comparison for ablation study of our ReshapeGAN on facial expression generation from RaFD dataset. The symbol $\uparrow$ ($\downarrow$) indicates that the larger (smaller) the value, the better the performance.}
\label{table:rafd_ablation}
\begin{tabular}{@{}P{1.35cm}M{.7cm}M{.7cm}M{.7cm}M{.7cm}M{.7cm}M{1.1cm}@{}}
\toprule
Method & SSIM$\uparrow$ & PSNR$\uparrow$ & MSE$\downarrow$ & RMSE$\downarrow$ & LPIPS$\downarrow$ & FID$\downarrow$\\
\midrule
Backbone & 0.9632 & 19.8962 & 0.0106 & 0.1017 & 0.1365 & 106.0801\\
Geometry & \textbf{0.9655} & 19.7940 & 0.0109 & 0.1030 & 0.1242 & 89.5522 \\
Geometry +$\mathcal{L}_{perturb}$ & 0.9636 & \textbf{20.1368} & \textbf{0.0101} & \textbf{0.0991} & 0.1241 & 81.6016\\
Geometry +$\mathcal{L}_{percep}$ & 0.9628 & 19.5127 & 0.0116 & 0.1064 & 0.1184 & 61.8619 \\
ReshapeGAN & 0.9641 & 20.1008 & 0.0103 & 0.0998 & \textbf{0.1030} & \textbf{49.2096}\\
\bottomrule
\end{tabular}
\end{table}

\subsubsection{Viewpoint transfer}

\begin{table}
\centering
\caption{Quantitative comparison of viewpoint transfer (facial pose translation) on FEI dataset. The symbol $\uparrow$ ($\downarrow$) indicates that the larger (smaller) the value, the better the performance.}
\label{table:fei}
\begin{tabular}{@{}P{2cm}@{}M{.7cm}M{.7cm}M{.7cm}M{.7cm}M{.7cm}M{1.1cm}@{}}
\toprule
Method & SSIM$\uparrow$ & PSNR$\uparrow$ & MSE$\downarrow$ & RMSE$\downarrow$ & LPIPS$\downarrow$ & FID$\downarrow$\\
\midrule
Pix2pix~\cite{isola2017image} & 0.9274 & 16.9358 & 0.01619 & 0.1228 & 0.2236 & 79.3540 \\
DR-GAN~\cite{tran2018representation} & 0.9384 & 17.4995 & 0.01385 & 0.1138 & 0.2001 & 83.5483\\
$\text{PG}^2$~\cite{ma2017pose} & 0.9354 & 17.3366 & 0.01447 & 0.1160 & 0.1887 & \textbf{70.7044} \\
ReshapeGAN & \textbf{0.9578} &\textbf{19.2299} & \textbf{0.00970} & \textbf{0.0941} & \textbf{0.1721} & 73.5864 \\
\bottomrule
\end{tabular}
\end{table}

\begin{figure}[!ht]
\centering
\includegraphics[width=1\linewidth]{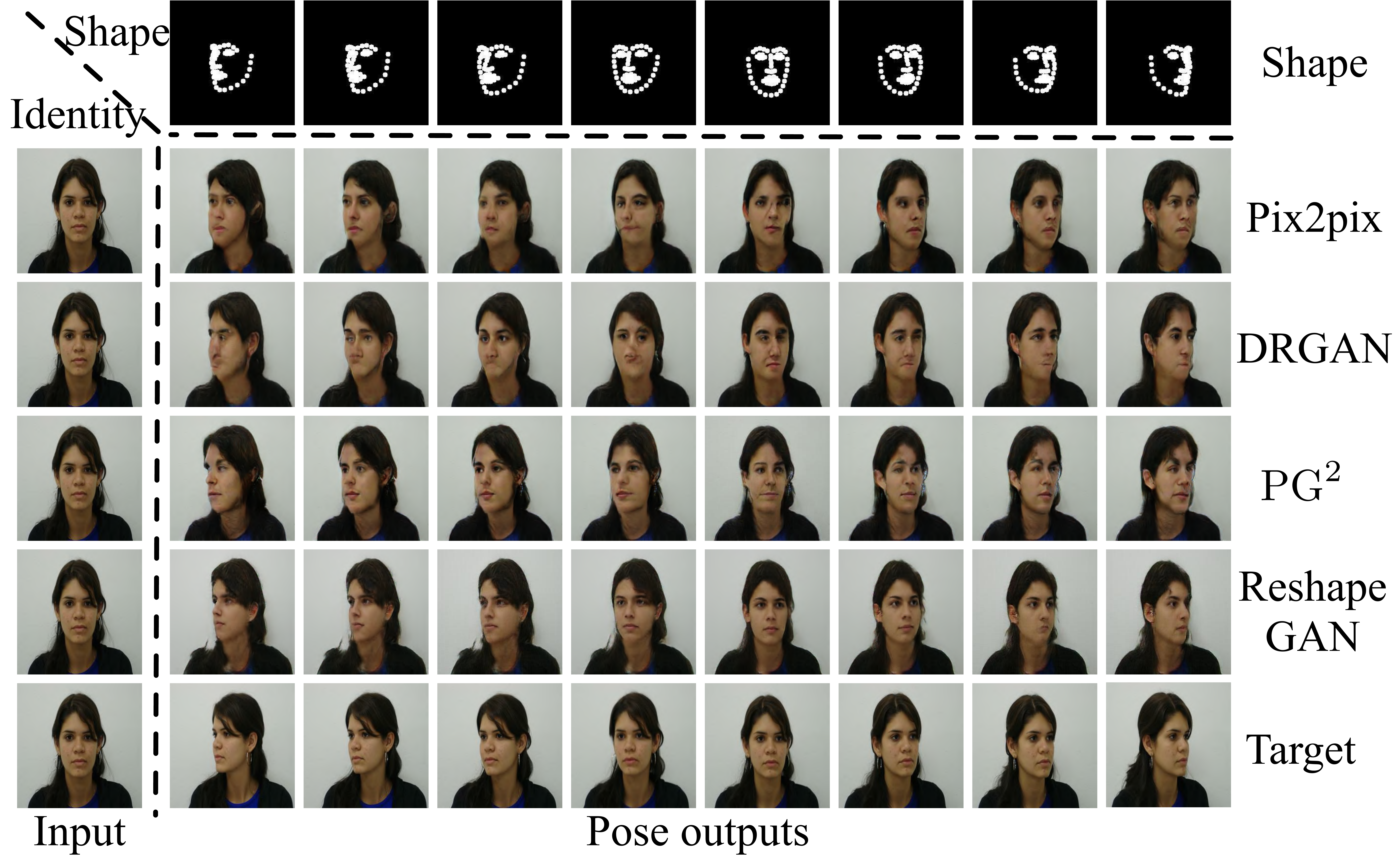}
\caption{Visual comparison of viewpoint transfer (facial pose translation) on FEI dataset.}
\label{fig:fei}
\end{figure}

Then, we devise viewpoint transfer task, aiming to generate different pose (viewpoint) faces on FEI dataset~\cite{thomaz2010new}, which contains 200 identities with 11 different pose directions. Since we can not extract the accurate geometric information from the full left or full right images, we don't use them for training. We use frontal images as inputs to generate other 8 pose images. We compare our ReshapeGAN with Pix2pix~\cite{isola2017image}, $\text{PG}^2$~\cite{ma2017pose} and DR-GAN~\cite{tran2018representation}. For ReshapeGAN, Pix2pix and $\text{PG}^2$, we also use the same settings as facial expression generation (Section~\ref{sec:facialexpression}) for adding guidance to these three methods, and for DR-GAN, we use the default setting of facial pose generation from the authors but using our training data. Besides, we use 160 identities for training and other 40 for testing. The quantitative comparison results are listed in Table~\ref{table:fei}. Our ReshapeGAN gets the lowest LPIPS and FID values among the four methods. Visual synthesized results are shown in Fig.~\ref{fig:fei} for comparison. Pix2pix generates images with blur boundary and some dirty color blocks, DR-GAN and $\text{PG}^2$ distort the faces, while our ReshapeGAN synthesizes better results with detailed parts (such as eyes, mouth and nose) as well as reasonable faces. By combining the geometric information, our ReshapeGAN can easily capture the relationships between different parts.

Note that, the geometric information does not limit to landmarks, instance maps and pose skeletons. We can also extract any other accurate geometric information and regard them as additional guidance to obtain better results.

\subsection{Reshaping by within-domain guidance with unpaired data}
\subsubsection{Controllable face translation on CelebA}

For this task, we aim to achieve the controllable facial reshaping using unpaired data. We conduct the experiments on CelebA dataset~\cite{liu2015faceattributes}, which contains approximately 200 thousand images with high diversity. To get precise facial expression, we reorganize this dataset for getting a sub-dataset using dlib~\cite{king2009dlib} to achieve face detection and crop the detected face regions then resize them to $256\times 256$. 
We use 157,619 images for training and 39,404 images for testing. In theory, our method could generate $39404\times 39404$ images on testing stage. Considering the huge consumption of computing resources, we only generate 1,000 images with different geometric representation for one input sample.

\begin{figure}[!ht]
\centering
\includegraphics[width=1\linewidth]{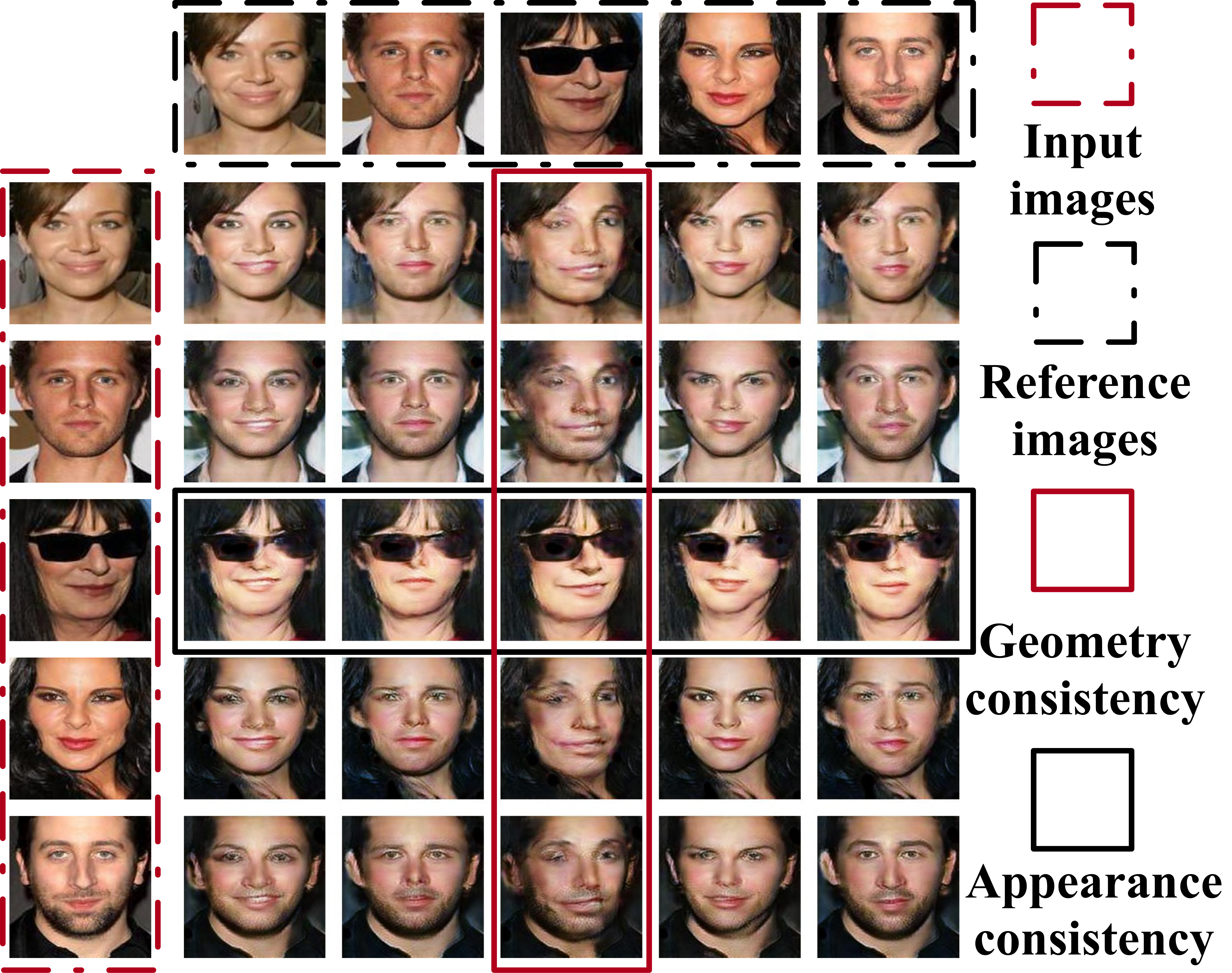}
\caption{The $5\times5$ outputs by our ReshapeGAN using random 5 images as both inputs and references on CelebA dataset. Images framed by dotted lines are input images and references images, and each row shows images with the identical appearance (e.g., the row framed by black line) while each column exhibits images with same geometric representation (e.g., the column framed by red line).}
\label{fig:celeba_demo}
\end{figure}

\begin{figure}[!ht]
\centering
\includegraphics[width=1\linewidth]{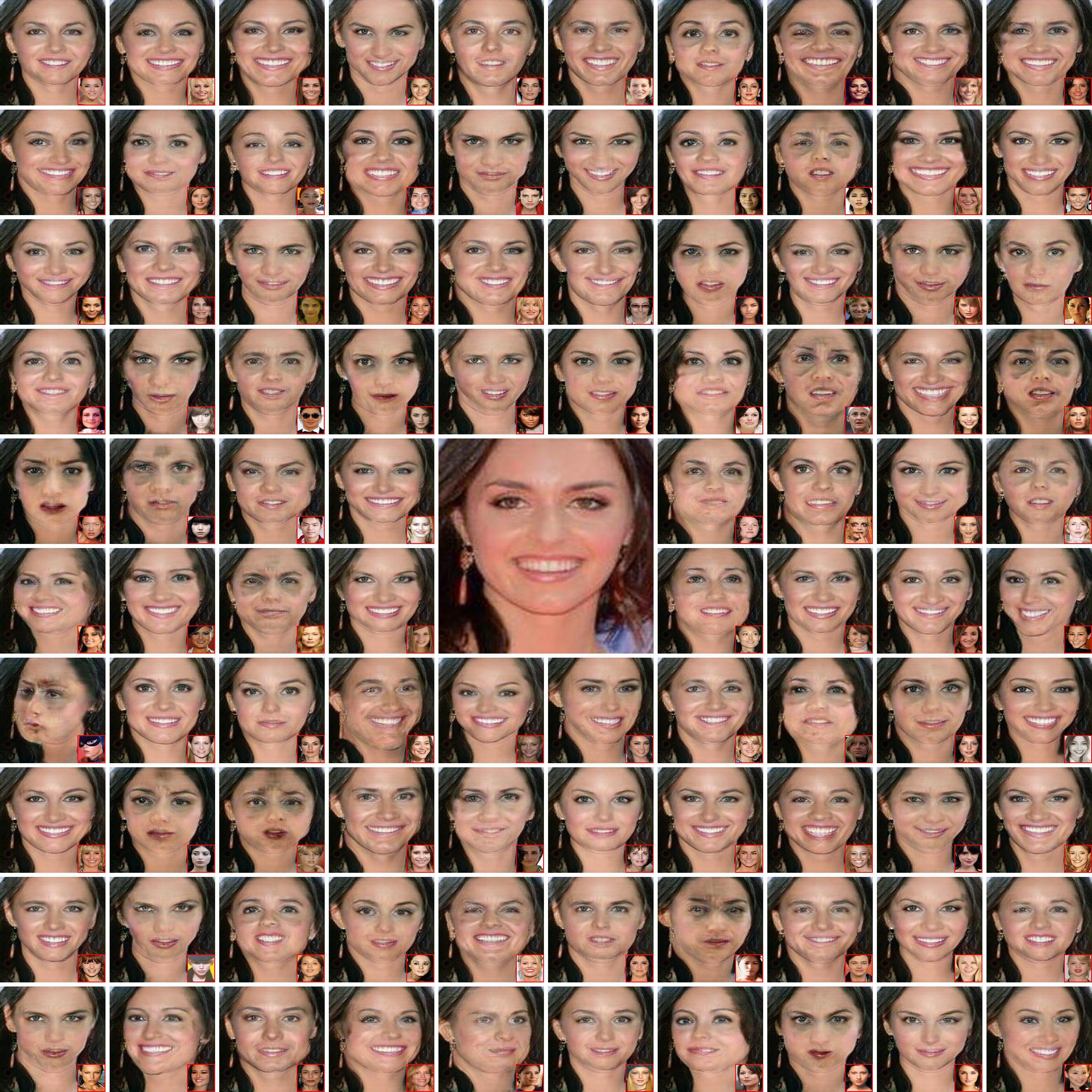}
\caption{Random 96 synthesized images by our ReshapeGAN from one input sample on CelebA dataset. The middle enlarged image shows the source input image. The smaller images framed by red box in the lower right corner of every generated image is the given reference image. The synthesized images have the same emotional and pose information with given reference images while preserving the appearance and identity information of input images. Please zoom in to see more details.}
\label{fig:celeba_100}
\end{figure}

Fig.~\ref{fig:celeba_demo} shows that our ReshapeGAN can synthesize 25 plausible images, via 5 different input images, where each row has the appearance and content consistency, while each column has geometric consistency. It can also be seen that, ReshapeGAN can generate images with required geometric representation by providing a geometric guidance, even if there are no paired samples for training, indicating that ReshapeGAN learns the facial expressions from the entire dataset rather than some specific sample. Moreover, Fig.~\ref{fig:celeba_100} exhibits the random 96 generated images using different geometric guidance. Note the smaller image framed by red box in the lower right corner of every generated image is the given reference image (for providing geometric information). The synthesized images preserve the appearance and identity of input images, and our ReshapeGAN is able to generate corresponding outputs with emotional and layout/pose information according to the given reference images. 

To our knowledge, we are the first to achieve arbitrary identity reshaping using unpaired data, with only limited images from each identity. Some traditional methods could perform face swapping, while Deepfakes\footnote{\url{https://github.com/deepfakes/Faceswap}} only translates one identity to another identity using abundant images from both source identity and target identity, but it fails to translate between multiple identities. Thus, we choose Faceswap from OpenCV\footnote{\url{https://opencv.org}}) as traditional method for comparison. Besides, we also make comparison using some popular image-to-image translation methods. Due to the lack of paired data, we conduct experiments using Pix2pix and $\text{PG}^2$ by removing the pixel-level loss between outputs and ground truth (denote as Pix2pix\texttt{--} and $\text{PG}^2$\texttt{--}) respectively. Here we regard the geometric information as the conditional information and provide the geometric information to generator by concatenating the raw input and geometry, through this way, we conduct experiments using StarGAN~\cite{choi2018stargan}, and in consideration of that there is only one domain in CelebA dataset, we remove the classification loss of StarGAN (denote as StarGAN\texttt{--}).

The quantitative comparison is given in Table~\ref{table:celeba}. SSIM and LPIPS scores computed by landmarks show the geometric consistency, higher SSIM and lower LPIPS scores indicate better geometric consistency with reference images. FID computes the distance between generated sample distribution and real reference images, and lower FID score represents better image generation quality. Our ReshapeGAN performs best in terms of SSIM, LPIPS and FID. We also use average identification distance to evaluate the identity similarity, Faceswap fails to preserve the identity with distance larger than 0.6, and the other methods can keep the identity information while our ReshapeGAN performs better on reshaping with larger distance (but below 0.6). The user study of identity/shape judgement can also arrive at similar conclusion that our ReshapeGAN works well considering both identity preservation and object reshaping.

\begin{figure}[!ht]
\centering
\includegraphics[width=1\linewidth]{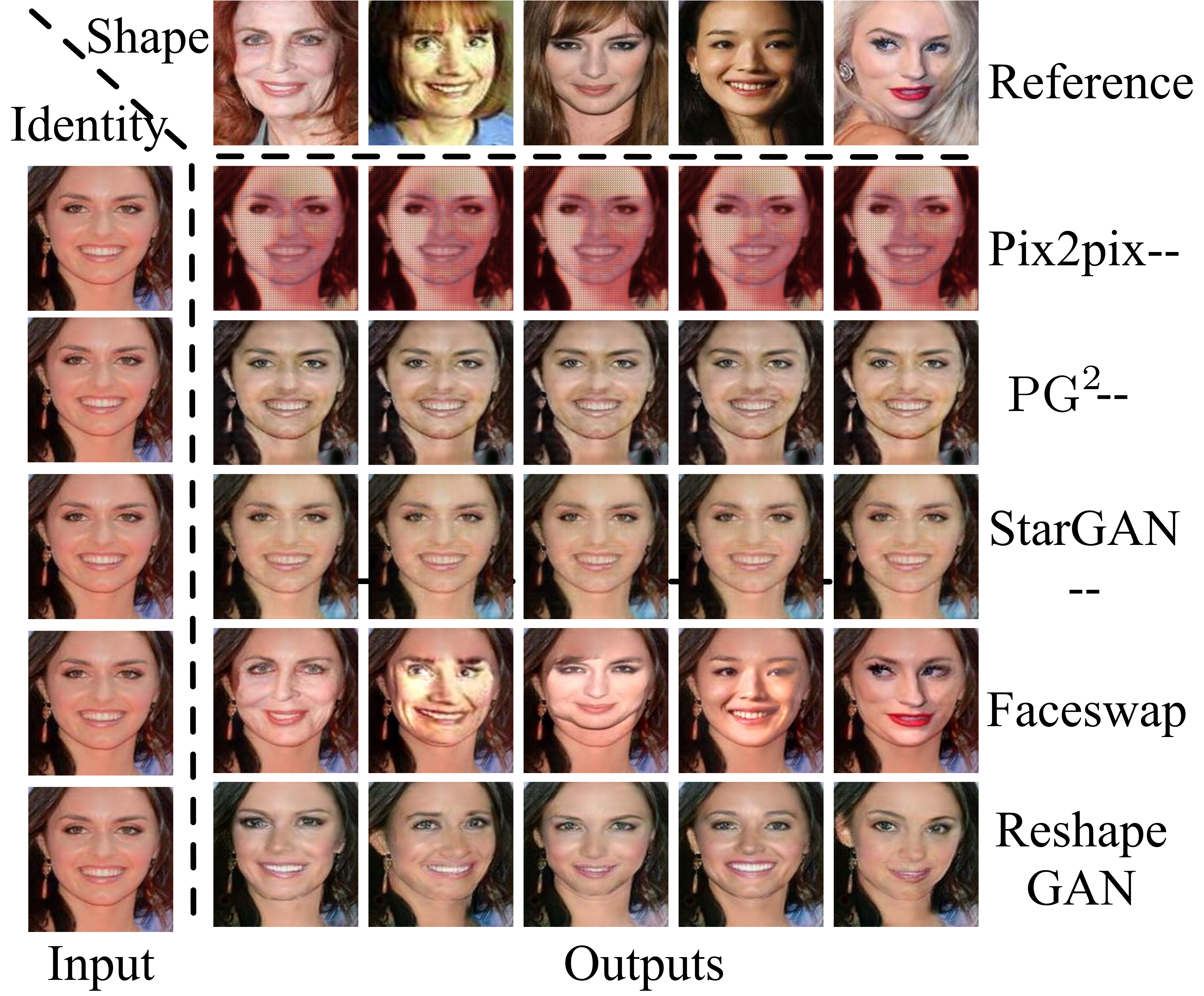}
\caption{Visual comparison of controllable face translation using different methods on CelebA dataset.}
\label{fig:Faceswap}
\end{figure}

\begin{figure}[!ht]
\centering
\includegraphics[width=1\linewidth]{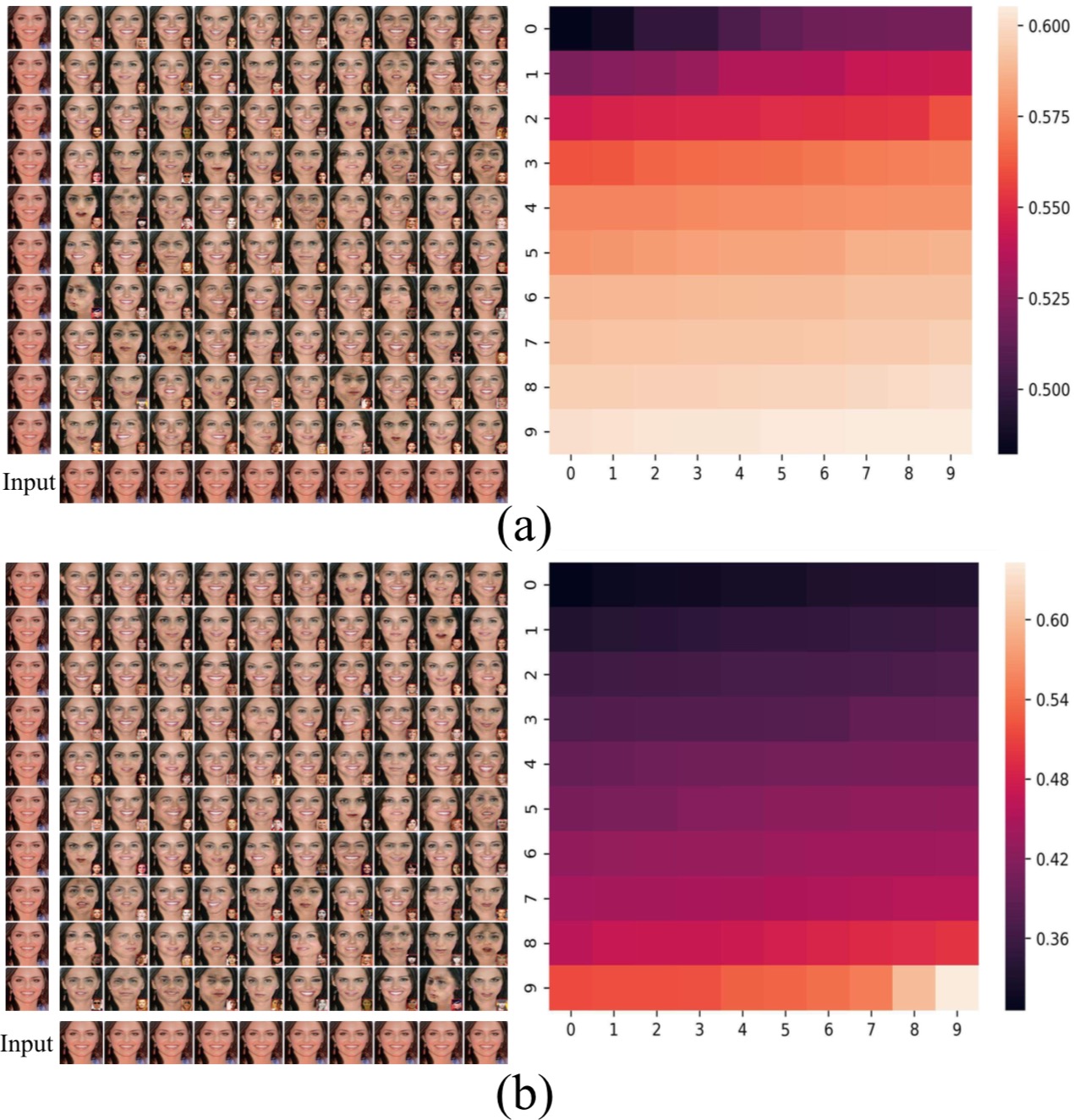}
\caption{(a) The random 100 generated results (left) sorted by progressive increase of identification distance (right) between the generated image and the source input image on CelebA dataset. (b) The random 100 generated results (left) sorted by progressive increase of LPIPS distance (right) between the corresponding target reference image and the source input image on CelebA dataset. We can see that the generated image with lower identification distance has higher similarity, while the generated image with lower LPIPS distance has higher quality. Please zoom in to see more details.}
\label{fig:celeba_100_visual}
\end{figure}

Visual comparison of different methods can be found in Fig.~\ref{fig:Faceswap}. As shown, Faceswap is limited to only borrow the face appearance from target reference and merging it with the corresponding region without considering the relationship between parts, and the output images seem implausible and don't preserve the identity information of the inputs. Pix2pix\texttt{--} and $\text{PG}^2$\texttt{--} fail to reshape the input images and generate images with artifacts. And StarGAN\texttt{--} also fails to synthesize images with required shape. Compared to these methods, our ReshapeGAN achieves face reshaping by providing one single reference image. The visual comparison of Fig.~\ref{fig:Faceswap} provides the same clues as quantitative comparison shown in Table.~\ref{table:celeba}. To show the powerful performance of our ReshapeGAN, we exhibit more synthesized results in Fig.~\ref{fig:celeba_100_visual}. We organize these generated results sorted by progressive increase of identification distance with source input image in Fig.~\ref{fig:celeba_100_visual}(a), and the right image show the visualization results of identification distance, which shows the relationship between the distance and the identity preservation with object reshaping.

\begin{table*}[!ht]
\centering
\caption{Quantitative comparison of controllable face translation on CelebA dataset. Higher SSIM and lower LPIPS scores represent better geometric matching with guided geometric information. Lower FID score means better image quality. Identification distance larger than 0.6 shows different identities, while larger distance below 0.6 indicates better reshaping. Higher votes to both identity and shape of user study indicate better reshaping performance. Please refer to Section~\ref{sec:evaluation} for details about evaluation metrics.}
\label{table:celeba}
\begin{tabular}{ccccccc}
\toprule
Method &  SSIM (Landmark) & LPIPS (Landmark) & FID & Identification Distance & Identity / Shape (User Study) \\
\midrule
Faceswap &  0.6949 & 0.1775 & 111.7921 & \underline{0.7230} & 0.025 / 0.605\\
Pix2pix\texttt{--} & 0.6049 & 0.2561 & 338.4090 & 0.4419& 0.771 / 0.093\\
$\text{PG}^2$\texttt{--} & 0.6605 & 0.2097 & 194.1435 & 0.3369 &  1.0 / 0.102\\
StarGAN\texttt{--} &  0.6544 & 0.2156 & 157.4540 & 0.2576 & 1.0 / 0.075\\
ReshapeGAN &  0.8332 & 0.0932 & 110.7313 &0.5700& 0.385 / 0.719\\
\bottomrule
\end{tabular}
\end{table*}

Based on the above synthesized results, we see that not all of the generated images can borrow the effective information reasonably, some generated images have extremely exaggerated regions. More interestingly, if the geometric information of given reference image is from a man while the input image depicts a woman, there is a chance that some sort of masculine pattern will be generated. We guess that the provided geometric information may somehow carry the gender and age characteristics. If the pose distance between input image and given reference image is too large, our model fails to generate plausible outputs. To obtain more intuitive results, we compute the LPIPS distance between each reference image and the input image, and visualize the results in Fig.~\ref{fig:celeba_100_visual}(b). Here we reorganize the generated images sorted by progressive increase of LPIPS distance between target reference image and source input image. As it can be seen, ReshapeGAN performs well if the LPIPS is less than an implicit threshold. However, when the LPIPS distance is too large, ReshapeGAN is not able to translate the input image to the given target geometric space with limited prior information.

\begin{table*}[!ht]
\centering
\caption{Quantitative comparison of controllable face translation on UTKFace dataset. Higher SSIM and lower LPIPS scores represent better geometric matching with guided geometric information. Lower FID score means higher image quality. Identification distance larger than 0.6 shows different identities, while larger distance below 0.6 indicates better reshaping. Higher votes to both identity and shape of user study indicate better reshaping performance. Please refer to Section~\ref{sec:evaluation} for details about evaluation metrics.}
\label{table:utk}
\begin{tabular}{cccccc}
\toprule
Method &  SSIM (Landmark) & LPIPS (Landmark) & FID &Identification Distance & Identity / Shape (User Study)\\
\midrule
Faceswap &  0.5595 & 0.2404 & 120.09001 &\underline{0.7354} &0.045 / 0.807\\
Pix2pix\texttt{--} &  0.6063 & 0.2197 & 251.4699 & 0.4144  &0.786 / 0.188\\
$\text{PG}^2$\texttt{--} & 0.5197 & 0.3127 & 173.5819 & 0.2628 & 1.0 / 0.024\\
StarGAN\texttt{--} &  0.5501 & 0.2793 & 186.4931 & 0.1850 &0.991 / 0.048\\
ReshapeGAN &  0.8103 & 0.0849 & 103.8411 &0.5902 &0.494 / 0.850\\
\bottomrule
\end{tabular}
\end{table*}
\subsubsection{Controllable face translation on UTKFace}
\begin{figure}[!ht]
\centering
\includegraphics[width=1\linewidth]{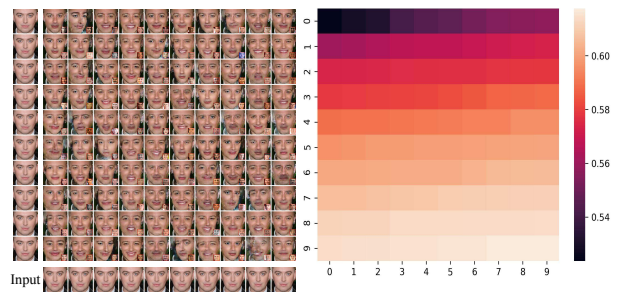}
\caption{The random 100 generated results (left) sorted by progressive increase of identification distance (right) between the generated image and the source input image on UTKFace dataset. Please zoom in to see more details.}
\label{fig:age}
\end{figure}

Then, to evaluate whether the geometric information carries underlying information such as gender and age characteristics, we conduct the face translation experiments on another large dataset UTKFace~\cite{zhang2017age}. This dataset contains over 20,000 face images with long age span (range from 0 to 116 years old). In this task, we also take experiments to generate images with arbitrary geometric information. According to our above assumption, the geometric information may contain underlying scale, position and facial expression information. By providing information to the generator and discriminator, they can build a mapping function between input images and geometric information. Visual results are shown in Fig.~\ref{fig:age}, we can see that the generated images have a large range of ages. We also visualize the identification distance, from which, we can see that the given geometric information does covers intrinsic characteristics. In order to fool the discriminator, the generator should dig in the implicit information and express it, reasonably.

\begin{figure}[!ht]
\centering
\includegraphics[width=1\linewidth]{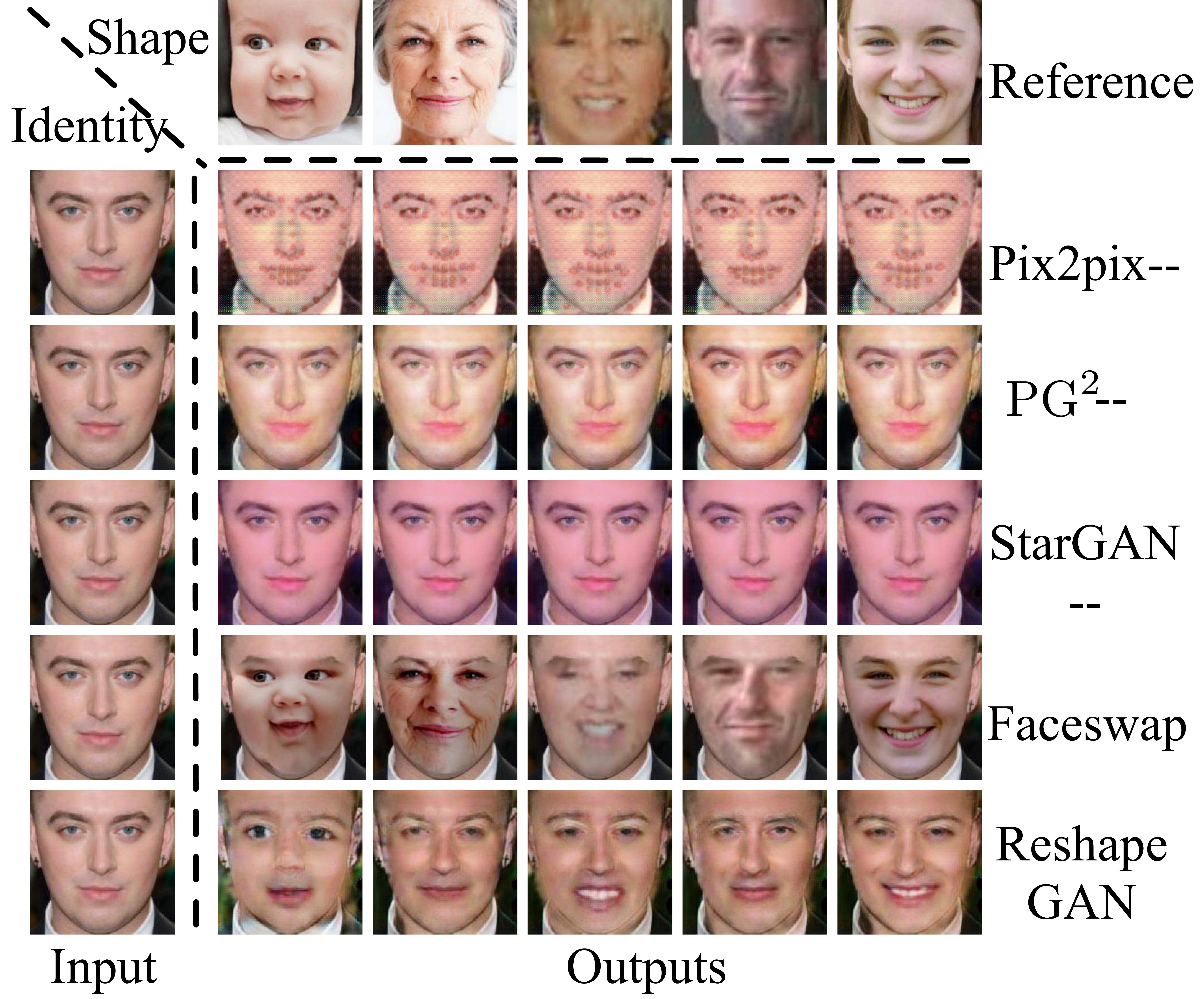}
\caption{Visual comparison of controllable face translation using different methods on UTKFace dataset.}
\label{fig:utk_compare}
\end{figure}

For this task, we also conduct some comparative experiments using aforementioned methods. The comparative visual results are shown in Fig.~\ref{fig:utk_compare}. We can see that the synthesized images using Pix2pix\texttt{--}, $\text{PG}^2$\texttt{--} and StarGAN\texttt{--} do not achieve the facial reshaping. Comparing with Faceswap, our ReshapeGAN can preserve the identity information better. Quantitative comparison of different methods are given in Table~\ref{table:utk}, as can be seen, our ReshapeGAN performs best with highest SSIM, lowest LPIPS and FID, indicating best geometric matching with highest image quality. Besides, ReshapeGAN also gets largest identification distance compared to all other methods below 0.6. Meanwhile, the reshaping ability with identity preservation of ReshapeGAN can also be validated by user study. 

\subsubsection{Controllable cat reshaping}
\begin{figure}[!ht]
\centering
\includegraphics[width=1\linewidth]{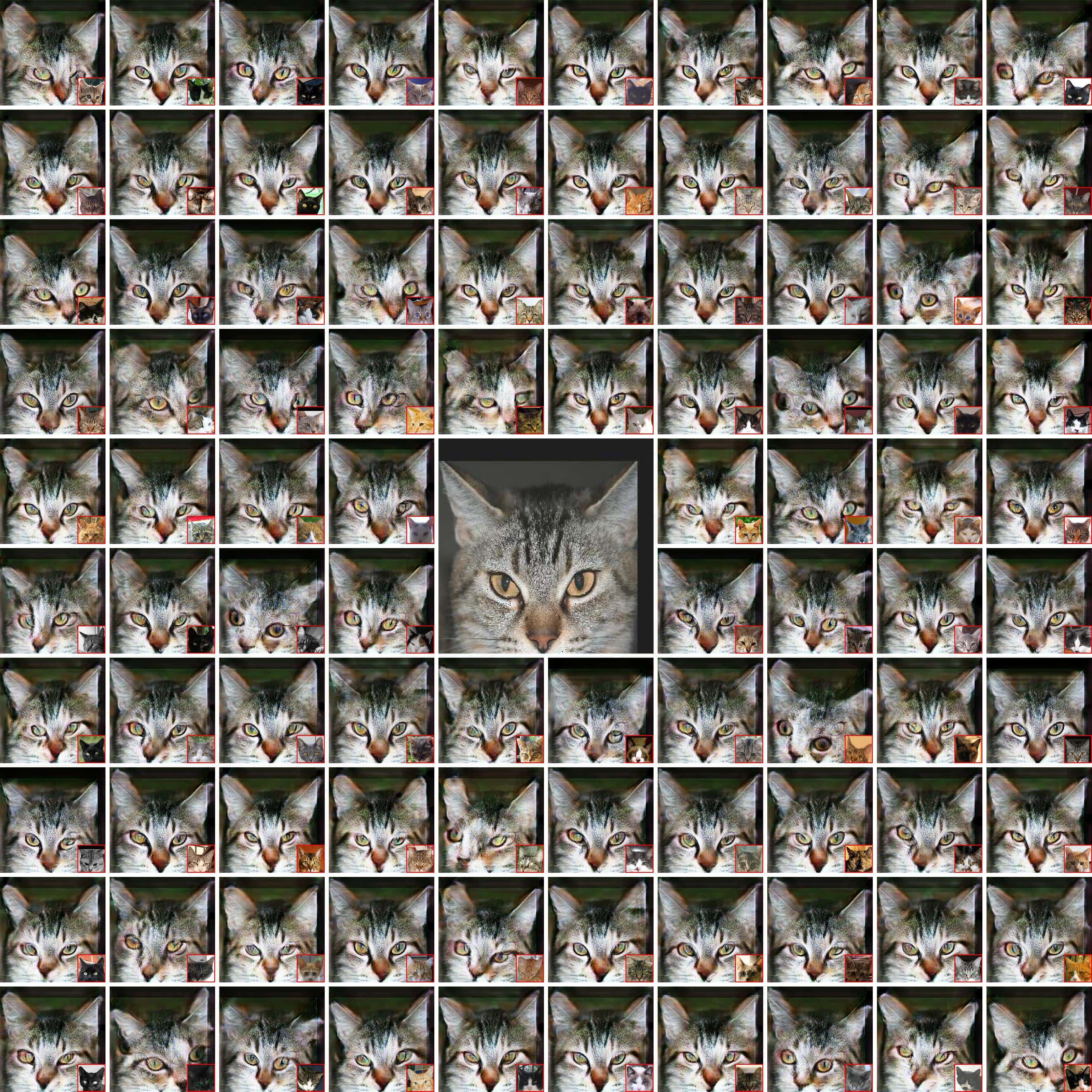}
\caption{Random 96 synthesized images by our ReshapeGAN from one input sample on cat head dataset. The middle enlarged image shows the source input image. And the smaller image framed by red box in the lower right corner of every generated image is the given target reference image. Please zoom in to see more details.}
\label{fig:cat}
\end{figure}

Moreover, we implement our ReshapeGAN on an interesting application that is to reshape a cat by providing geometric information. We use data and annotations from the cat head dataset~\cite{zhang2008cat}, and each annotation file describes the coordinates of 9 defined points (6 for ears, 2 for eyes and 1 for mouth). Here we use 7,287 cropped images (according to the coordinates) for training and 1,000 images for testing. Experimental results are shown in Fig.~\ref{fig:cat}, our ReshapeGAN can reshape the cat image with a given geometric information, while keeping the detailed information of input image. For this task, we only provide the limited 9 pointed annotation, which describes the location of ears, eyes and mouth. So, without sufficient guidance, our method can also work well. For this task, we only compute the FID between generated cat images and real images for reference, and the FID score is 76.7446.

\subsubsection{Controllable pose reshaping}
\begin{figure}[!ht]
\centering
\includegraphics[width=1\linewidth]{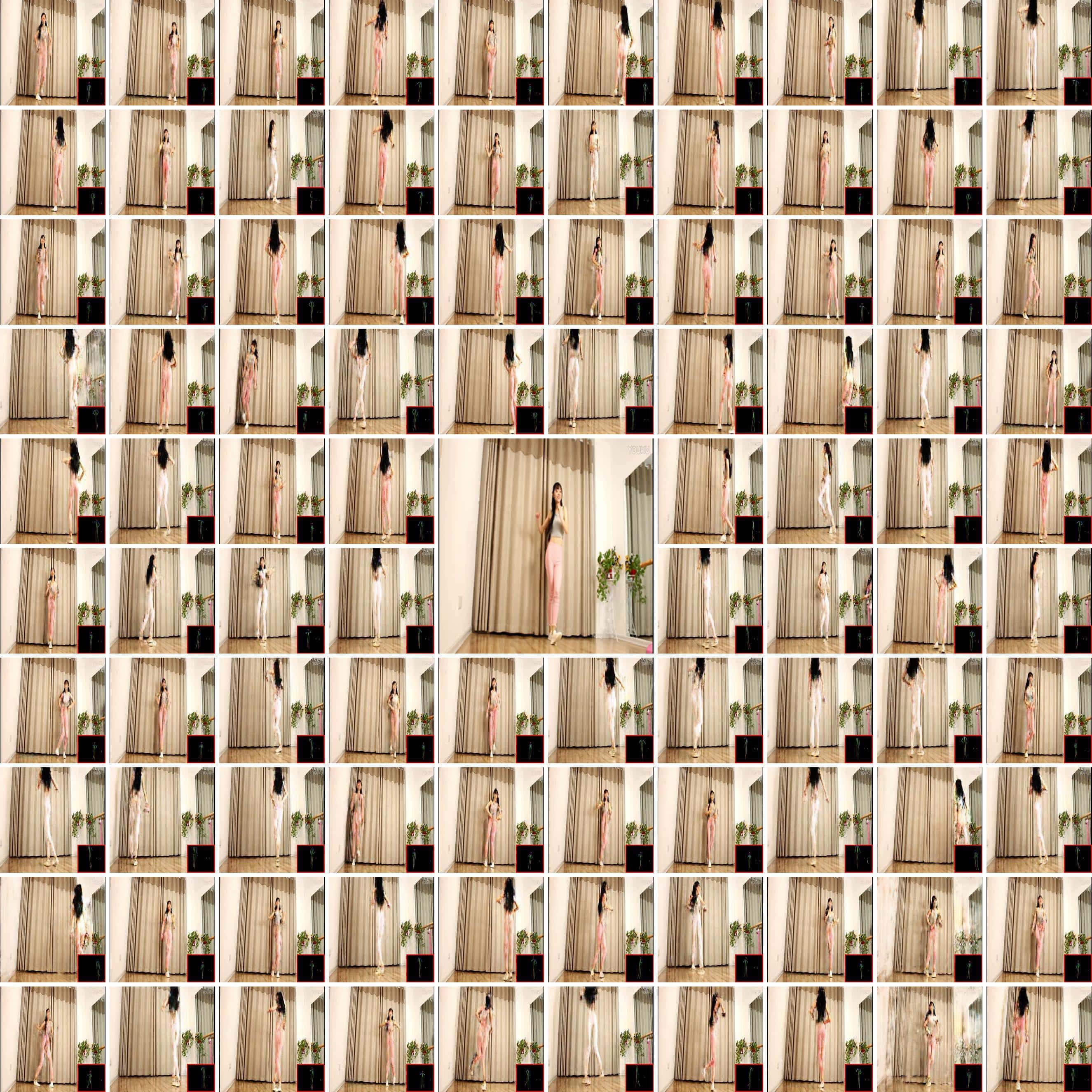}
\caption{Random 96 synthesized images by our ReshapeGAN from one input sample on Panama dataset. The middle enlarged image shows the source input image. And the smaller image framed by red box in the lower right corner of every generated image is the given pose reference image extracted using OpenPose. We could generate a different video sequence with poses from target identity while using appearance and identity information from source inputs. Please zoom in to see more details.}
\label{fig:pose}
\end{figure}

In addition, the skeleton guidance of human body provides more detailed pose information. Thus, we use pose estimation model OpenPose~\cite{cao2017realtime} to extract pose information of human images and regard them as geometric guidance for pose reshaping application. Here we perform our ReshapeGAN on Panama dataset\footnote{\url{https://github.com/llSourcell/Everybody_Dance_Now}}. Note that, we don't use the paired training for this task, and only add the cycle-consistency constrains to our model. By fusing pose information and appearance information from given reference images, our method can generate reasonable results as shown in Fig.~\ref{fig:pose}. Here we only generate images by frames, we leave the pose sequence generation by adding spatial and temporal constrains as our future work. For this task, we also only compute the FID between generated pose images and real images for reference, and the FID score is 145.3365.

\subsubsection{Ablation study}

\begin{table*}[!ht]
\centering
\caption{Quantitative comparison of different cases for our ReshapeGAN without or with geometric information on CelebA dataset. Only adding geometric information to generator $G$ (Only G) and discriminator $D$ (Only D) get very low identification distance, showing that they fail to achieve geometric reshaping with given geometric constrain, that is, they fall into the trivial solution to generate the very similar output to the input image (very lower identification distance). The user study also validates the same point (higher votes to both identity and shape of user study indicate better reshaping performance). Our ReshapeGAN, by adding geometric information to both $G$ and $D$, gets higher SSIM and lower LPIPS scores, indicating to reshape the input image with given geometric information better, and also gets the lower FID score representing higher image quality.}
\label{table:archi}
\begin{tabular}{cccccc}
\toprule
Method & SSIM (Landmark) & LPIPS (Landmark) & FID &Identification Distance&  Identity / Shape (User Study)\\
\midrule
Only G &  0.6763 & 0.1855 & 168.2893 & 0.0438 &1.0 / 0.1739\\
Only D &  0.6543 & 0.1924 & 176.5256 & 0.0352 &1.0 / 0.073\\
G+D (ReshapeGAN) & 0.8332 & 0.0932 & 110.7313 & 0.5700 &0.385 / 0.719\\
\bottomrule
\end{tabular}
\end{table*}

\begin{table*}[!ht]
\centering
\caption{Quantitative comparison of different cases for our ReshapeGAN without or with perceptual loss on CelebA dataset. Although the model without the perceptual loss can get better performance in terms of FID, LPIPS and SSIM, the identification distance is larger than 0.6. Thus the perceptual loss does helps to preserve the identity and appearance information from source input image.}
\label{table:perceptual}
\begin{tabular}{ccccccc}
\toprule
Method &  SSIM (Landmark) & LPIPS (Landmark) & FID &Identification Distance& Identity / Shape (User Study)\\
\midrule
w/o $\mathcal{L}_{percep}$ & 0.8683 & 0.0706 & 84.6713 & \underline{0.6075} &0.145 / 0.879\\
w/ $\mathcal{L}_{percep}^{gr}$ &  0.8509 & 0.0774 & 99.9562 &\underline{0.6188} & 0.186 / 0.934\\
w/ $\mathcal{L}_{percep}^{gi}$ &  0.8332 & 0.0932 & 110.7313 &0.5700 &0.385 / 0.719\\
\bottomrule
\end{tabular}
\end{table*}

\begin{table*}[!ht]
\centering
\caption{Quantitative comparison of different cases for our ReshapeGAN without or with perturbed loss on CelebA dataset. The perturbed loss can reduce the identification distance and reserve the source identity information while helping improve image quality.}
\label{table:perturbed}
\begin{tabular}{cccccc}
\toprule
Method &  SSIM (Landmark) & LPIPS (Landmark) & FID &Identification Distance& Identity / Shape (User Study)\\
\midrule
w/o $\mathcal{L}_{perturb}$ & 0.8473 & 0.0827 & 115.4396 & 0.5790  & 0.256 / 0.896\\
W/ $\mathcal{L}_{perturb}$ & 0.8332 & 0.0932 & 110.7313 & 0.5700 & 0.385 / 0.719\\
\bottomrule
\end{tabular}
\end{table*}

\begin{figure*}[!ht]
\centering
\includegraphics[width=1.0\linewidth]{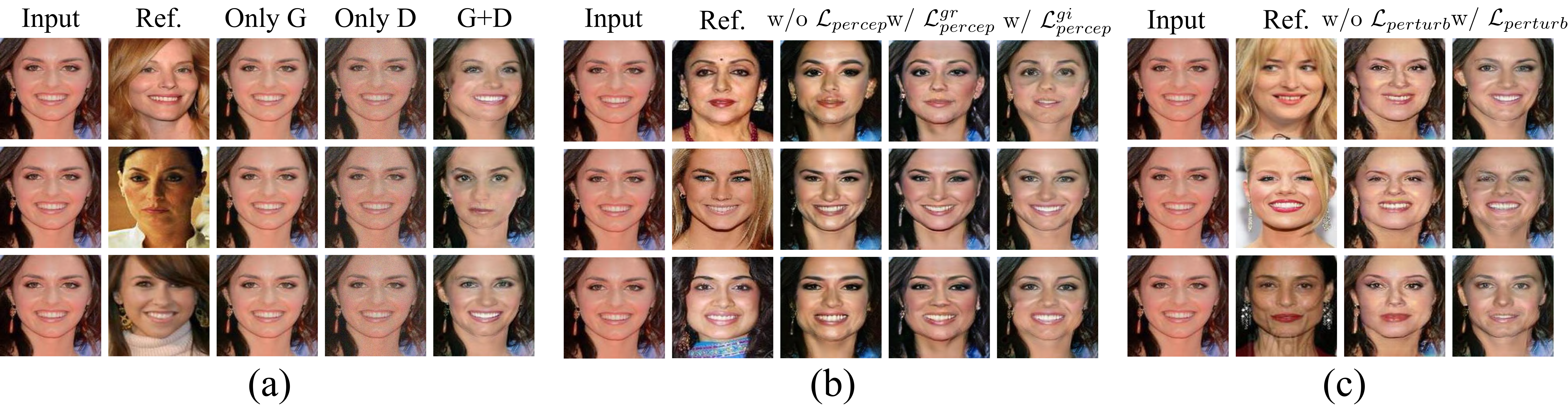}
\caption{Visual synthesized results of different cases without or with geometric information (a), of different cases without or with perceptual loss (b), of different cases without or with perturbed loss (c), on CelebA dataset.}
\label{fig:celeba_ablation}
\end{figure*}

On the one hand, to investigate the architecture of our geometry-guided method, we compare three cases for generator and discriminator without or with geometric information: 1) only add geometric information to generator, 2) only add geometric information to discriminator, 3) add geometric information to both generator and discriminator. We conduct experiments on CelebA dataset for this ablation study in the three cases, and the visual results are shown in Fig.~\ref{fig:celeba_ablation}(a). Discriminator without geometric information guidance captures the pose information but fails to provide effective gradient direction to generator, so that the model can not generate required controllable images.  If we don't provide the given geometric information to generator, the model can not achieve controllable reshaping. Thus the guidance is necessary for both generator and discriminator. Table~\ref{table:archi} with quantitative comparison also concludes this point.

On the other hand, we explore the efficiency of perceptual loss by considering three cases: 1) without perceptual loss, 2) with perceptual loss computed between synthesized fake image and real reference image ($\mathcal{L}_{percep}^{gr}$), 3) with perceptual loss computed between synthesized fake image and real input image ($\mathcal{L}_{percep}^{gi}$). Visual results can be found in Fig.~\ref{fig:celeba_ablation}(b), and quantitative results are listed in Table~\ref{table:perceptual}. From the results, we can see that the model falls in trivial solution and encounters over-fitting problem by adding constrains between generated images and given target reference images, that is, the generated images are very similar to the given reference images. Meanwhile, computing the perceptual loss between synthesized images and input images can guarantee content and identity consistency. Comparing with L1 loss, perceptual loss focuses on high-level semantic matching between images rather than pixel-level matching. 

Furthermore, we also devise similar experiments to explore the effectiveness of the perturbed loss, the results are shown in Table~\ref{table:perturbed} and Fig.~\ref{fig:celeba_ablation}(c). As shown, the perturbed loss helps to preserve the identity information and improves the generation quality. Although the model without perturbed loss can get a little higher SSIM and lower LPIPS scores (computed between geometric information), we should pay more attention to preserve the appearance consistency and image generation quality.

\subsection{Reshaping by cross-domain guidance}

\subsubsection{Facial reshaping across multiple datasets}
In this task, we use 5 facial image datasets including Yale~\cite{georghiades1997yale}, WSEFEP~\cite{olszanowski2015warsaw}, ADFES~\cite{van2011moving}, KDEF~\cite{calvo2008facial} and RaFD~\cite{rafdcite} for facial reshaping. 

\begin{figure}[!ht]
\centering
\includegraphics[width=1\linewidth]{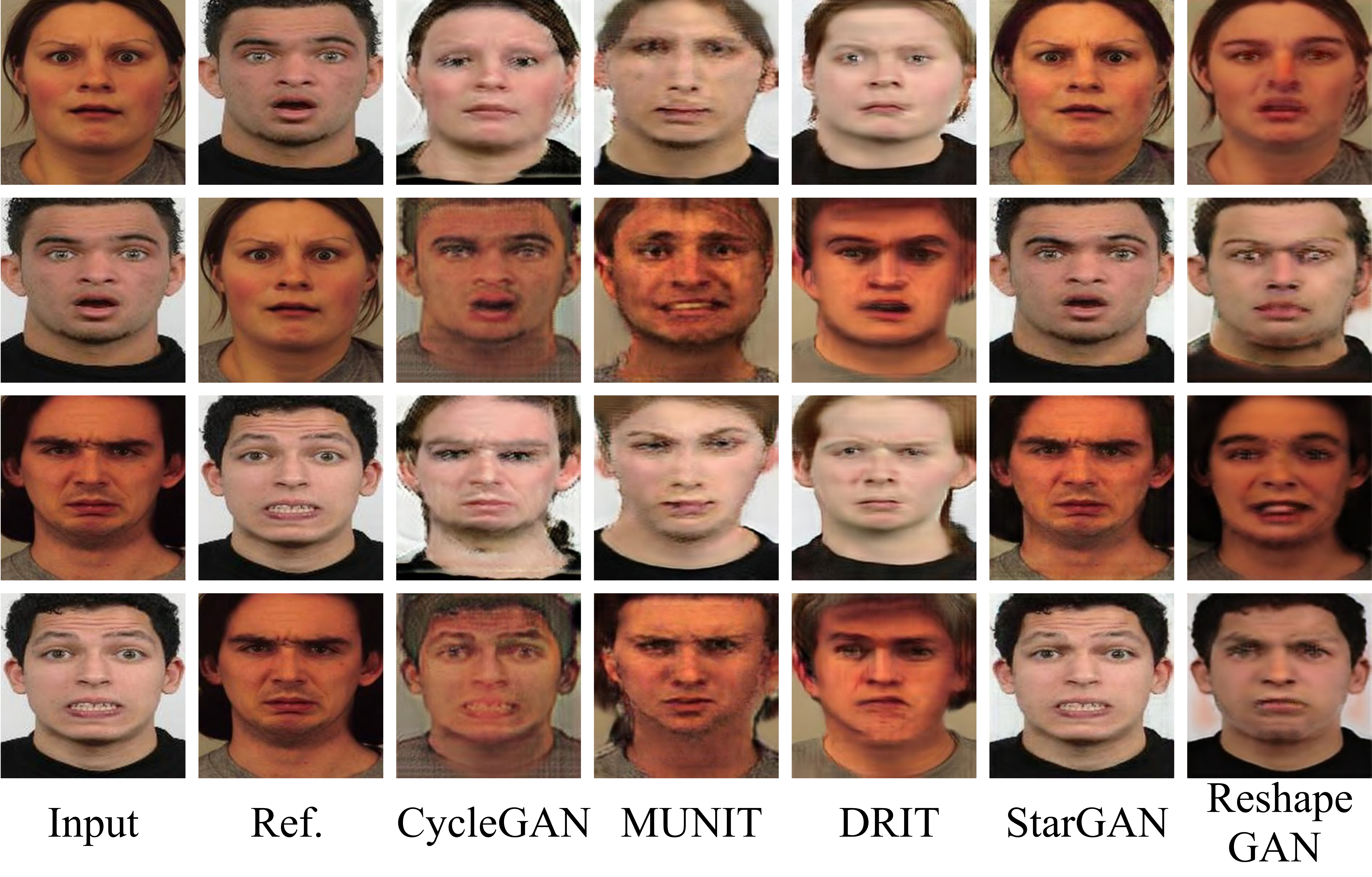}
\caption{Visual comparison of reshaping by cross-domain guidance for using different methods on KDEF-RaFD dataset.}
\label{fig:kdef_rafd}
\end{figure}

\begin{figure}[!ht]
\centering
\includegraphics[width=1.0\linewidth]{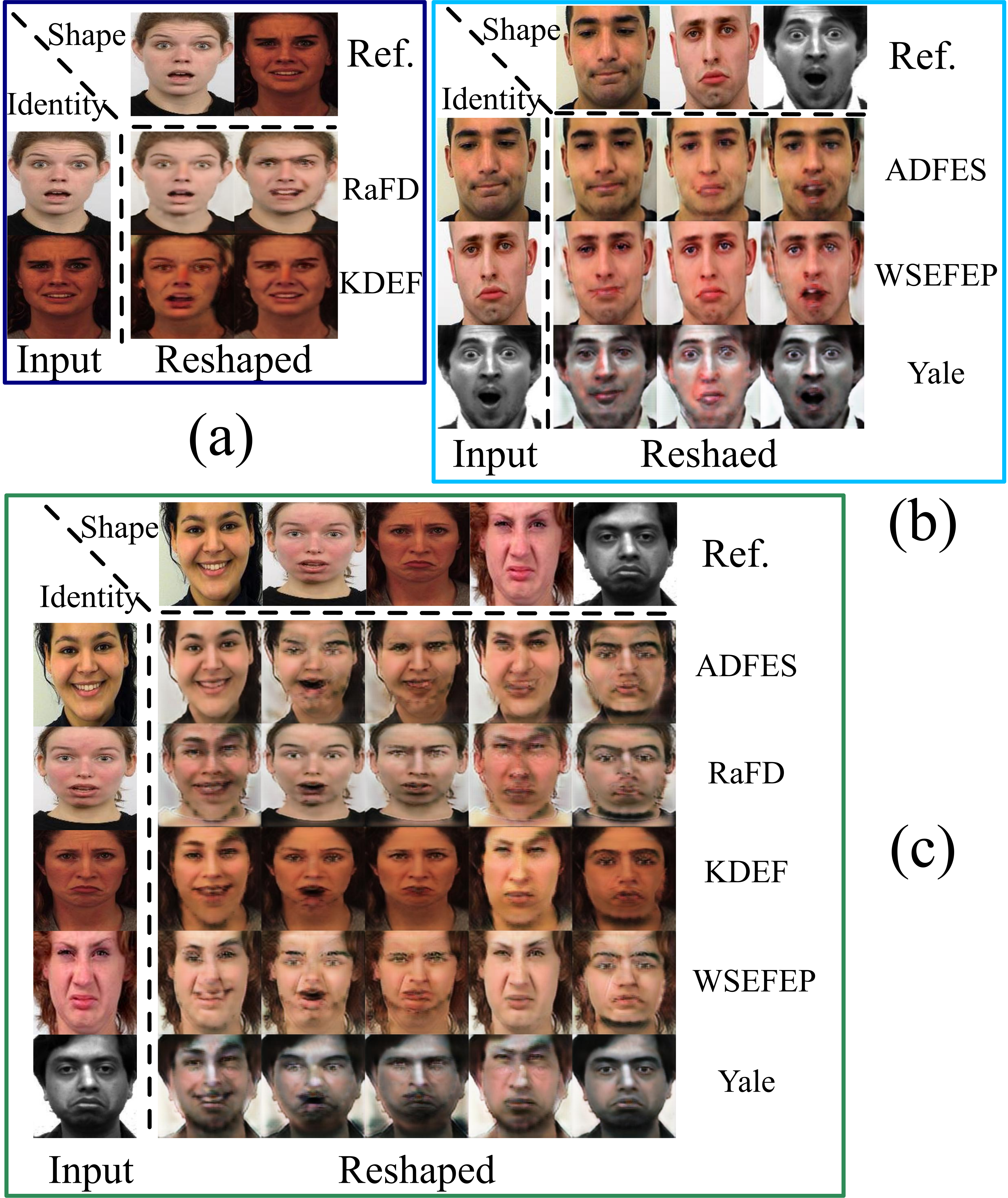}
\caption{Visual results synthesized by our ReshapeGAN on RaFD and KDEF dataset (a), on WSEFEP, ADFES, and Yale (b), on above five datasets (c), for facial reshaping.}
\label{fig:facial}
\end{figure}

\begin{table*}[!ht]
\centering
\caption{Quantitative comparison of controllable face translation on KDEF-RaFD dataset. Higher SSIM and lower LPIPS scores represent better geometric matching with guided geometric information. Lower FID score means higher image quality. Identification distance larger than 0.6 shows different identities, while larger distance below 0.6 indicates better reshaping. Please refer to Section~\ref{sec:evaluation} for details about evaluation metrics. The results above the dashed line are computed in the first case: the input image is from KDEF and the reference image is from RaFD, while the results below the dashed line are computed in the second case: the input image is from RaFD and the reference image is from KDEF.}
\label{table:rafd_kdef}
\begin{tabular}{cccccc}
\toprule
Method  & SSIM (Landmark) & LPIPS (Landmark) & FID& Identification Distance \\
\midrule
CycleGAN~\cite{zhu2017unpaired} & 0.6181 & 0.2608 & 172.8081 & 0.4591\\
MUNIT~\cite{huang2018multimodal}   & 0.6083 & 0.2754 & 182.4521  & 0.4721\\
DRIT~\cite{lee2018diverse}     & 0.6949 & 0.2692 & 115.8099 & 0.5427\\
StarGAN~\cite{choi2018stargan}   & 0.6623  & 0.2493 &  130.7385& 0.0612\\
ReshapeGAN  & 0.7981 & 0.1184 & 100.8653 & 0.5909 \\
\hdashline
CycleGAN~\cite{zhu2017unpaired}   & 0.6323 & 0.2372 & 140.2661& 0.4602 \\
MUNIT~\cite{huang2018multimodal}    & 0.6315 & 0.2482 & 154.2534& 0.4654  \\
DRIT~\cite{lee2018diverse}     & 0.6521 & 0.1932 & 84.2044 & \underline{0.6361} \\
StarGAN~\cite{choi2018stargan}   & 0.6256 & 0.2503 &  100.9369& 0.0489\\
ReshapeGAN  & 0.7895 & 0.1222 & 122.9286  & 0.5930\\
\bottomrule
\end{tabular}
\end{table*}

First, considering domain gaps between different datasets, we devise three different experiments for evaluating the efficiency of our ReshapeGAN. First, we conduct experiments on RaFD and KDEF datasets, and we choose CycleGAN, MUNIT, DRIT and StarGAN for comparison. In order to compare fairly, we provide the geometric information to all the generators as the additional constrains for all compared methods. Additionally, for StarGAN, we use the one-hot encoding to perform the facial translation on the two different datasets. By combining all the emotional categories appearing on both two datasets, we get 7 different emotional labels (contemptuous, disgusted, fearful, happy, sad, and surprised) and conduct experiments following~\cite{choi2018stargan}. For the testing stage, we randomly select 91 input-reference pairs from the testing images of the two domains. Please note that there are two cases for the cross-domain reshaping on KDEF-RaFD datasets: 1) the input image is from KDEF and the reference is from RaFD, 2) the input image is from RaFD and the reference image is from KDEF. The visual synthesized images are exhibited in Fig.~\ref{fig:kdef_rafd} and Fig.~\ref{fig:facial}(a). CycleGAN and MUNIT only achieve the domain adaption according to the given reference image, and they fail to perform reshaping by combining the geometric information. DRIT generates images with underlying position changes, but it also fails to reshape input images with given reference images. Above all, recent unpaired cross-domain methods among two different domains (CycleGAN, MUNIT and DRIT) can achieve domain adaption (including the low-level texture and background translation), but they pay more attention to the low-level matching rather than the high-level geometrical matching, that is, they fail to generate images with required shape by providing the geometric guidance. StarGAN fails to achieve the facial translation by providing additional emotional label, we guess the reason is that the same two datasets contain similar emotional expression but different background information and data distribution, so that the model is hard to focus on the emotional representation. Compared to above methods, our ReshapeGAN can achieve the facial reshaping by only providing a single reference image. For the quantitative comparison, Table~\ref{table:rafd_kdef} lists all the results. As shown, our method gets highest SSIM and lowest LPIPS scores, which indicates that ReshapeGAN achieves geometric consistency with reference images. At the same time, our method can preserve the appearance information (including the background and domain information) well in terms of identification distance.

Second, we perform our method on WSEFEP, ADFES and Yale datasets, where the three datasets have small intra-domain distances, and the visual results are shown in Fig.~\ref{fig:facial}(b). For the above two cases, our method can perform well on both object reshaping and generation quality.

Finally, we conduct experiments on all above five emotional datasets, and the visual results can be seen in Fig.~\ref{fig:facial}(c). We observe that the synthesized images have dirty color blocks if the domain gap is large (such as KDEF and WSEFEP), since the samples from the two datasets have extremely different location as well as pose information, such that our method can not generate reasonable outputs without sufficient prior information.

\subsubsection{Caricature reshaping across multiple datasets}
\begin{figure}[!ht]
\centering
\includegraphics[width=1\linewidth]{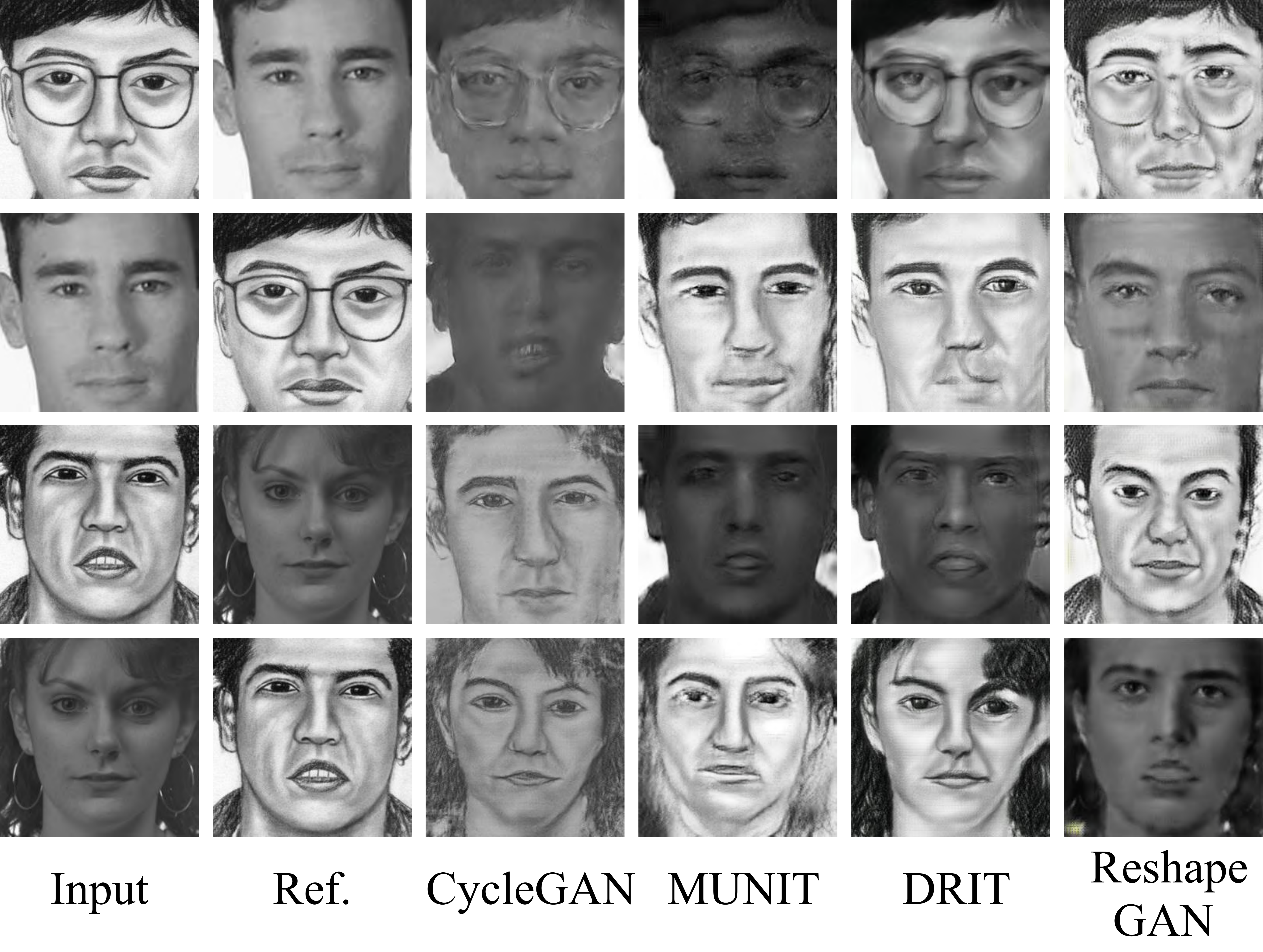}
\caption{Visual comparison of reshaping by cross-domain guidance for using different methods on PHOTO-SKETCH dataset.}
\label{fig:photo_sketch}
\end{figure}

For this task, we aim to achieve caricature translation between multiple datasets. The datasets include CUHK Face Sketch database~\cite{wang2008face}, KDEF~\cite{calvo2008facial}, IIIT-CFW~\cite{mishra2016iiit} and PHOTO-SKETCH~\cite{wang2009face}. Among the four datasets, we could get 5 different image style domains (there are 2 domains from PHOTO-SKETCH dataset). For each dataset, we randomly select approximately four fifths of samples for training and others for testing. First, we conduct comparative experiments on PHOTO-SKETCH dataset. Due to the lack of conditional label, here we don't use StarGAN for comparison and follow the training/testing splitting in~\cite{yi2017dualgan}. Using the same testing criteria, we randomly get 199 input-reference images. Note that the input-reference paired images are not from the same identity. The visual synthesized results can be seen in Fig.~\ref{fig:photo_sketch}. As shown, our ReshapeGAN can reshape the input images according to the shape of reference images, while CycleGAN, MUNIT and DRIT only achieve the domain transfer between two different image manifolds, failing to find the geometric mapping function. The quantitative comparison is listed in Table~\ref{table:photo_sketch}, from which, our ReshapeGAN can keep better geometric consistency with effective identification distance. Although CycleGAN, MUNIT and DRIT have lower FID scores by achieving domain adaption, they all actually fail to reshape the input images according to the reference images and preserve the appearance and style information of input images. 

\begin{table*}[!ht]
\centering
\caption{Quantitative comparison of controllable face translation on PHOTO-SKETCH dataset. Higher SSIM and lower LPIPS scores represent better geometric matching with guided geometric information. Lower FID score means higher image quality. Identification distance larger than 0.6 shows different identities, while larger distance below 0.6 indicates better reshaping. Please refer to Section~\ref{sec:evaluation} for details about evaluation metrics. The results above the dashed line are computed in the first case: the input image is from PHOTO domain and the reference image is from SkETCH domain, while the results below the dashed line are computed in the second case: the input image is from SKETCH domain and the reference image is from PHOTO domain.}
\label{table:photo_sketch}
\begin{tabular}{ccccc}
\toprule
Method & SSIM (Landmark) & LPIPS (Landmark) & FID & Identification Distance \\
\midrule
CycleGAN~\cite{zhu2017unpaired}   & 0.4839 & 0.3567 & 45.6227 & \underline{0.6320}\\
MUNIT~\cite{huang2018multimodal}   & 0.4623 & 0.3462 & 89.6343  & \underline{0.6103}\\
DRIT~\cite{lee2018diverse}     & 0.4823 & 0.3514 & 67.8397 & \underline{0.6420} \\
ReshapeGAN  & 0.8088 & 0.0871 & 125.4597  & 0.5749\\
\hdashline
CycleGAN~\cite{zhu2017unpaired}   & 0.4786 & 0.3703 & 77.2058 & \underline{0.6038} \\
MUNIT~\cite{huang2018multimodal}    & 0.4823 & 0.3643 & 84.2434  & \underline{0.6243}\\
DRIT~\cite{lee2018diverse}     & 0.4927 & 0.3513 & 65.3231  & \underline{0.6182}\\
ReshapeGAN   & 0.7931 & 0.1123 & 101.5948 & 0.5818\\
\bottomrule
\end{tabular}
\end{table*}

\begin{figure}[!ht]
\centering
\includegraphics[width=1.0\linewidth]{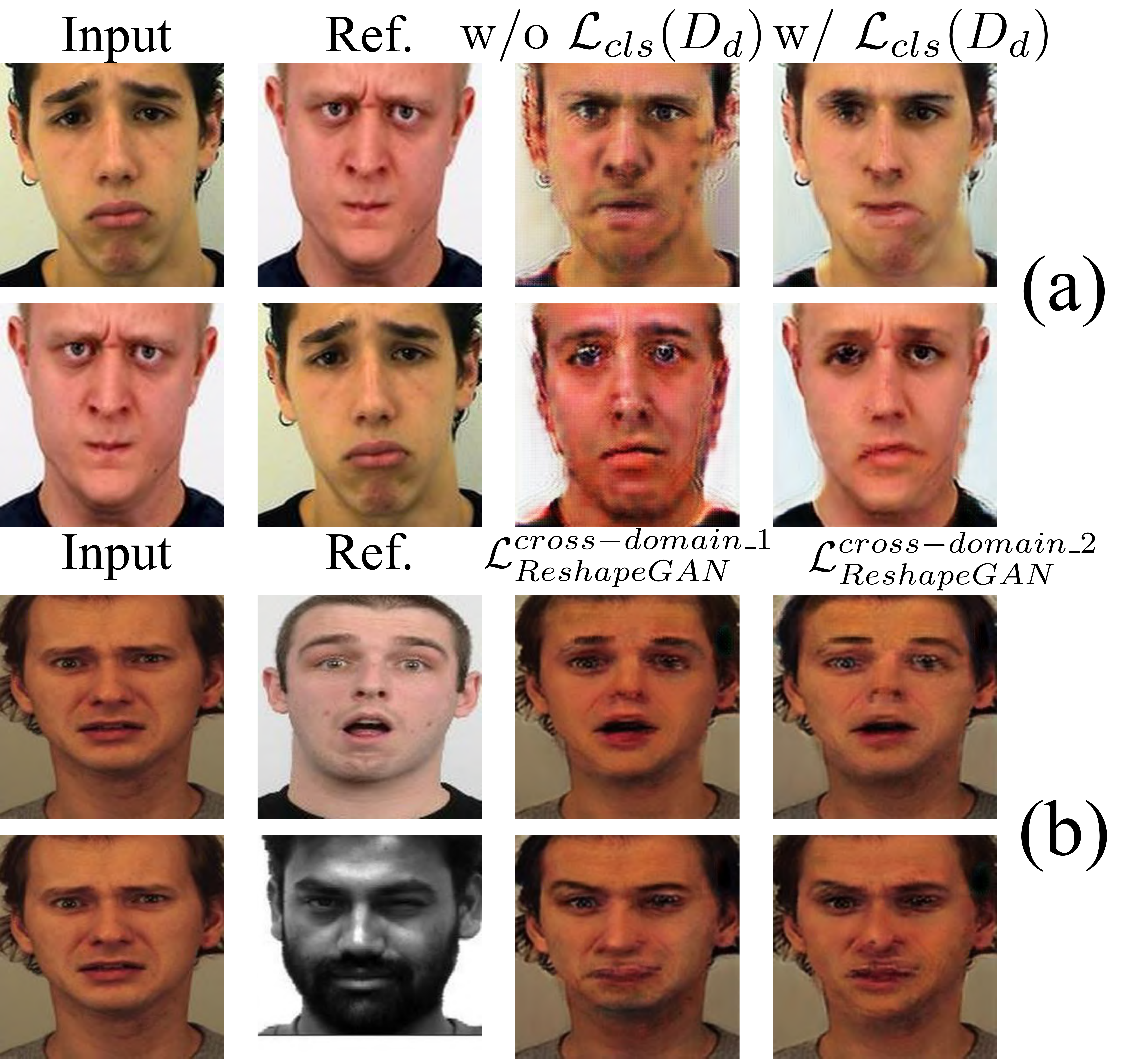}
\caption{Visual comparison results generated by our ReshapeGAN of different cases without or with $\mathcal{L}_{cls}(D_{d})$ (a), of different cases with $\mathcal{L}^{cross-domain\_1}_{ReshapeGAN}(G,D,D_d)$ or with $\mathcal{L}^{cross-domain\_2}_{ReshapeGAN}(G,D,D_d)$ (b) on selected 5 facial datasets.}
\label{fig:ablation_cross}
\end{figure}

\begin{figure*}[!ht]
\centering
\includegraphics[width=.9\linewidth]{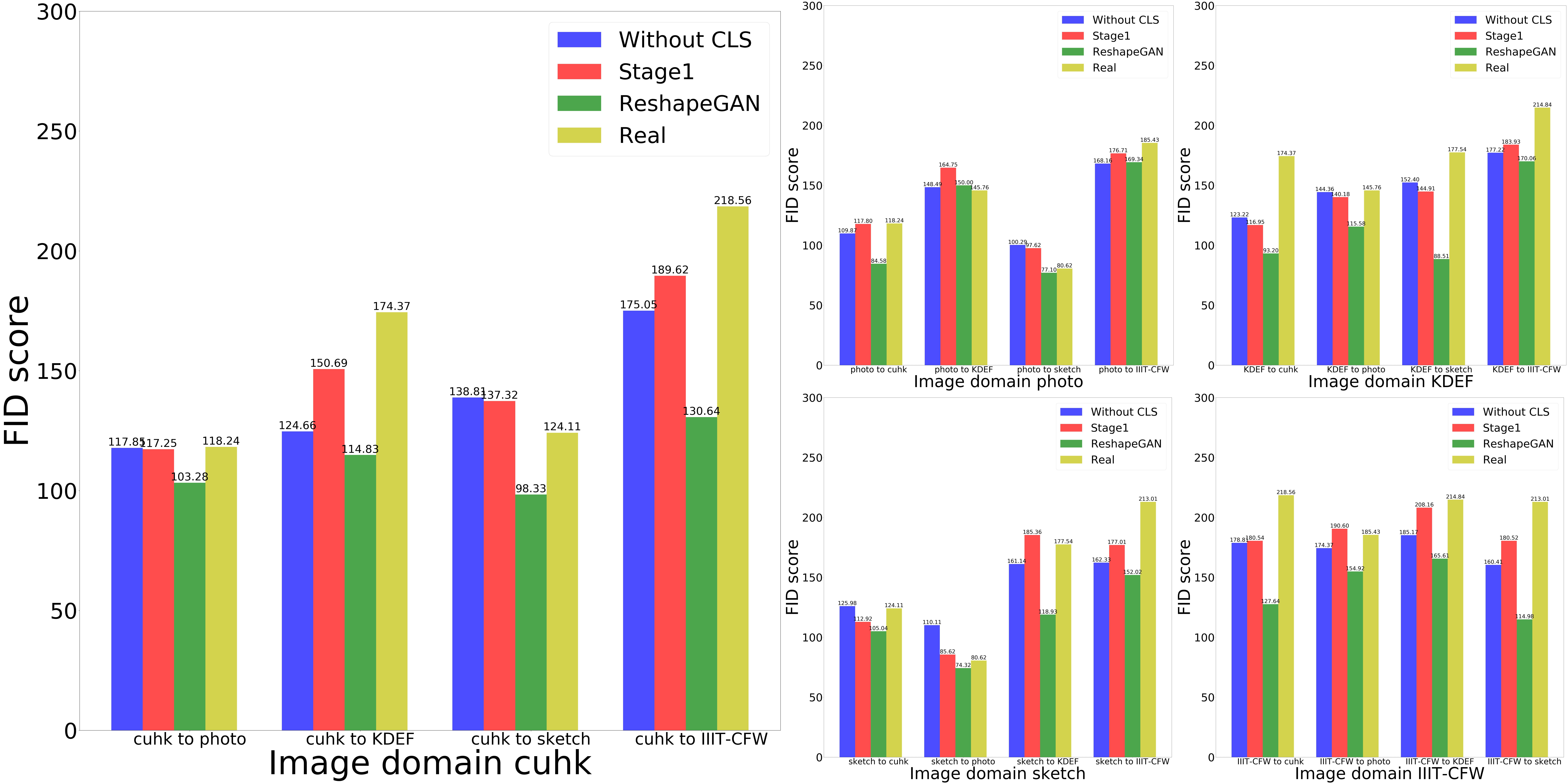}
\caption{The visualization results of of FID distance using different methods on four datasets for caricature reshaping.}
\label{fig:visual}
\end{figure*}

Actually, in the testing stage, we can generate arbitrary domain image with arbitrary geometric guidance, and we can also generate $n\times n$ images based on $n$ images from 5 image domains for this task. To reduce the computational cost, we randomly assemble 5 images from 5 image domains to obtain a combination. So we can get $n=\max{n_{i}}$ combinations, where $n_{i}$ represents the number of testing images from the 5 image domains separately. In order to evaluate the performance of our ReshapeGAN on cross-domain reshaping, we compute the FID scores between generated images and real images. Here we compute the FID scores of 4 different cases: 1) between synthesized images without the domain classification loss $\mathcal{L}_{cls}(D_{d})$ and reference images (Without CLS for abbreviation), 2) between synthesized images optimized by $\mathcal{L}^{cross-domain\_1}_{ReshapeGAN}(G,D,D_d)$ and reference images (Stage1 for abbreviation), 3) between synthesized images optimized by $\mathcal{L}^{cross-domain\_2}_{ReshapeGAN}(G,D,D_d)$ and reference images (ReshapeGAN for abbreviation), 4) between source input images and reference images (Real for abbreviation). We visualize all the FID results in Fig.~\ref{fig:visual}. Each sub figure shows the translation from a source domain to another 4 target domains. We see that the FID score of the model without domain classification loss is higher than the model with domain classification loss (ReshapeGAN). So the domain classification loss does helps to improve the image generation quality. Besides, if we split a difficult cross-domain object reshaping task to two easier tasks, we can get better generation performance. Almost all the FID scores of proposed model are lower than those of other models using only one stage. That is, the second refining stage does improve the image generation quality. More ablation study can be found in Section~\ref{multiple_ablation}. With geometric information from target domain, we see that the FID between generated images and reference images is lower than FID between source input images and reference images. Thus our ReshapeGAN can reduce domain distances by combining the geometric information.


\subsection{Ablation study}
\label{multiple_ablation}
In order to show the effectiveness of our pipeline architecture, we design another ablation study with one-stage generation architecture by combining domain classification loss $\mathcal{L}_{cls}(D_{d})$ and geometric information to a strong conditional input. Fig.~\ref{fig:ablation_cross}(a) shows the visual comparison results, from which, we can see that that the one-stage generation method without $\mathcal{L}_{cls}(D_{d})$ can not preserve the identity and domain information well, and the generation is not sharp while the boundary is not clear. And with $\mathcal{L}_{cls}(D_{d})$, the model generates images without dirty color blocks. In addition, we also explore the effectiveness of the second refining stage, the visual comparison is exhibited in Fig.~\ref{fig:ablation_cross}(b). As shown, the generated images with refining stage have better geometric changes according to given guidance. 

\section{Failure and limitation}

\begin{figure}[!ht]
\centering
\includegraphics[width=0.9\linewidth]{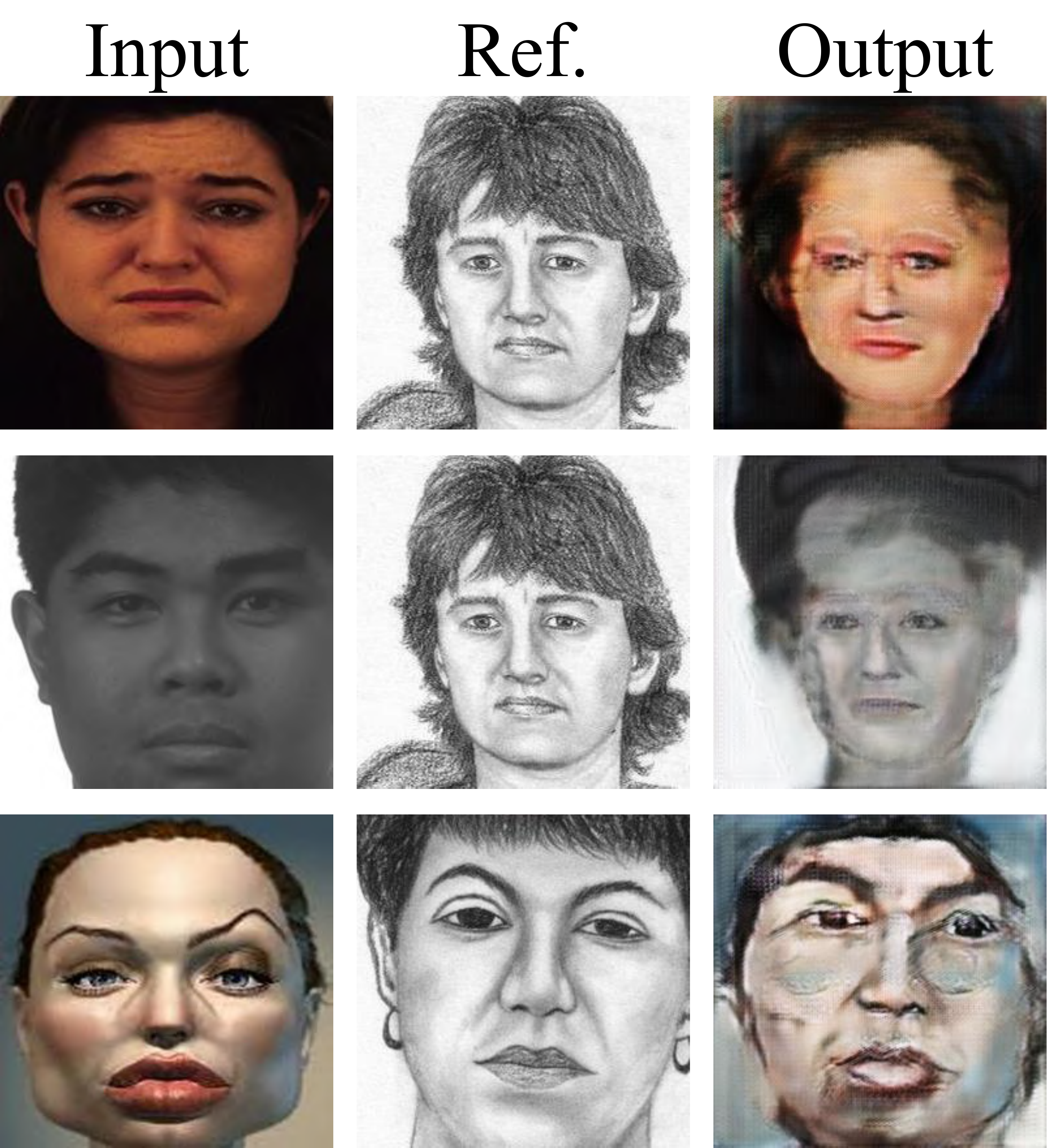}
\caption{The failure cases on object reshaping by our ReshapeGAN across multiple datasets.}
\label{fig:failure}
\end{figure}

In this section, we will discuss the limitation of our ReshapeGAN. Our model requires extra geometric information as an input, e.g., we use dlib~\cite{king2009dlib} to obtain face landmark as geometric information. So the face landmark should be exist and accurate, and sometimes we have to abandon the face images that dlib cannot extract landmarks (e.g., the method fails to obtain geometric information for the full left and right images on FEI dataset). Obviously, it will affect the performance of our model if the geometric information is wrong. For using the unsupervised manner in our experiments to achieve the cross-dataset image reshaping, we still lack prior knowledge when the semantic domain gap is huge between different image domains. Fig.~\ref{fig:failure} shows some failure cases of ReshapeGAN, where our model fails to preserve the appearance of input images well, and the outputs look like just the simple combination of source image and target image. Visually, our ReshapeGAN fails to reshape the input image to the reference geometric shape if the distance between the inputs and the references is too large, since our model has insufficient corresponding prior information to achieve reasonable translation. In this case, we will try to disentangle the content representation to make the content information preserved, and we leave this as our future work.

\section{Conclusion}
In this paper, we proposed a new architecture that can generate domain-specific or cross-domain images with geometric guidance, which is the first general framework for a wide range of object reshaping by providing a single reference. Comprehensive experiments on both ablation study and comparisons with comparable state-of-the-art models, in terms of all kinds of applicable quantitative and qualitative as well as human subjective evaluation metrics, show that our model performs better for identity preserved object reshaping.

\section*{Acknowledgment}
This work was supported in part by the Royal Society under IEC\textbackslash R3\textbackslash \ 170013 - International Exchanges 2017 Cost Share (Japan and Taiwan only), in part by a Microsoft Research Asia (MSRA) Collaborative Research 2019 Grant, and in part by the National Natural Science Foundation of China under grants 61771440 and 41776113.

\bibliographystyle{spmpsci}      

{\footnotesize
\bibliography{ObjReshape}
}


\end{document}